\documentclass[lettersize,journal]{IEEEtran}
\usepackage{amsmath,amsfonts}
\usepackage{algorithmic}
\usepackage{algorithm}
\usepackage{array}
\usepackage[caption=false,font=normalsize,labelfont=sf,textfont=sf]{subfig}
\usepackage{textcomp}
\usepackage{stfloats}
\usepackage{url}
\usepackage{verbatim}
\usepackage[pdftex]{graphicx}
\usepackage[pdftex]{color}
\usepackage{cite}
\usepackage{amssymb,amsopn,bm}
\usepackage{afterpage}
\usepackage{color}
\usepackage{url}
\usepackage{color}

\begin{document}

\title{Crowd-Powered Photo Enhancement \\ Featuring an Active Learning Based Local Filter}

\author{{Satoshi Kosugi and Toshihiko Yamasaki,\IEEEmembership{~Member,~IEEE}}
        % <-this % stops a space
% \thanks{This paper is supported by ???.}% <-this % stops a space
\thanks{The authors are with the Department of Information and Communication Engineering,
The University of Tokyo, Bunkyo-ku, Tokyo, Japan.

Copyright © 2023 IEEE. Personal use of this material is permitted. However, permission to use this material for any other purposes must be obtained from the IEEE by sending an email to pubs-permissions@ieee.org.}}

% The paper headers
\markboth{IEEE TRANSACTIONS ON CIRCUITS AND SYSTEMS FOR VIDEO TECHNOLOGY, VOL. XX, NO. XX, 2023}%
{Satoshi Kosugi and Toshihiko Yamasaki: Crowd-Powered Local Photo Enhancement Featuring Active Learning Based Parameter Optimization}

% \IEEEpubid{0000--0000/00\$00.00~\copyright~2021 IEEE}
% Remember, if you use this you must call \IEEEpubidadjcol in the second
% column for its text to clear the IEEEpubid mark.

\maketitle

\begin{abstract}
  In this study, we address local photo enhancement
	to improve the aesthetic quality of an input image
	by applying different effects to different regions.
	Existing photo enhancement methods are either not content-aware or not local; therefore,
	we propose a crowd-powered local enhancement method
	for content-aware local enhancement, which is achieved
	by asking crowd workers to locally optimize parameters for image editing functions.
	To make it easier to locally optimize the parameters,
	we propose an active learning based local filter.
	The parameters need to be determined at only a few key pixels selected by an active learning method,
	and the parameters at the other pixels are automatically predicted using a regression model.
	The parameters at the selected key pixels are independently optimized,
	breaking down the optimization problem into a sequence of single-slider adjustments.
	Our experiments show that the proposed filter outperforms existing filters,
	and our enhanced results are more visually pleasing than the results
	by the existing enhancement methods.
  Our source code and results are available at https://github.com/satoshi-kosugi/crowd-powered.

\end{abstract}


\section{Introduction}

\IEEEPARstart{P}{hotos} are taken daily,
but the quality may be low because of
poor lighting conditions~\cite{wei2018deep,chen2018learning,cai2018learning},
limited quality of cameras~\cite{ignatov2017dslr},
and inexperienced retouching skills~\cite{bychkovsky2011learning}.
To retouch such photos automatically,
this study addresses photo enhancement.
Global enhancement, which gives the same effect to an entire image,
cannot properly enhance images where the brightness is non-uniform.
To apply different effects to different regions,
we propose a local photo enhancement method.

\begin{figure}[t]
  \includegraphics[width=1\hsize]{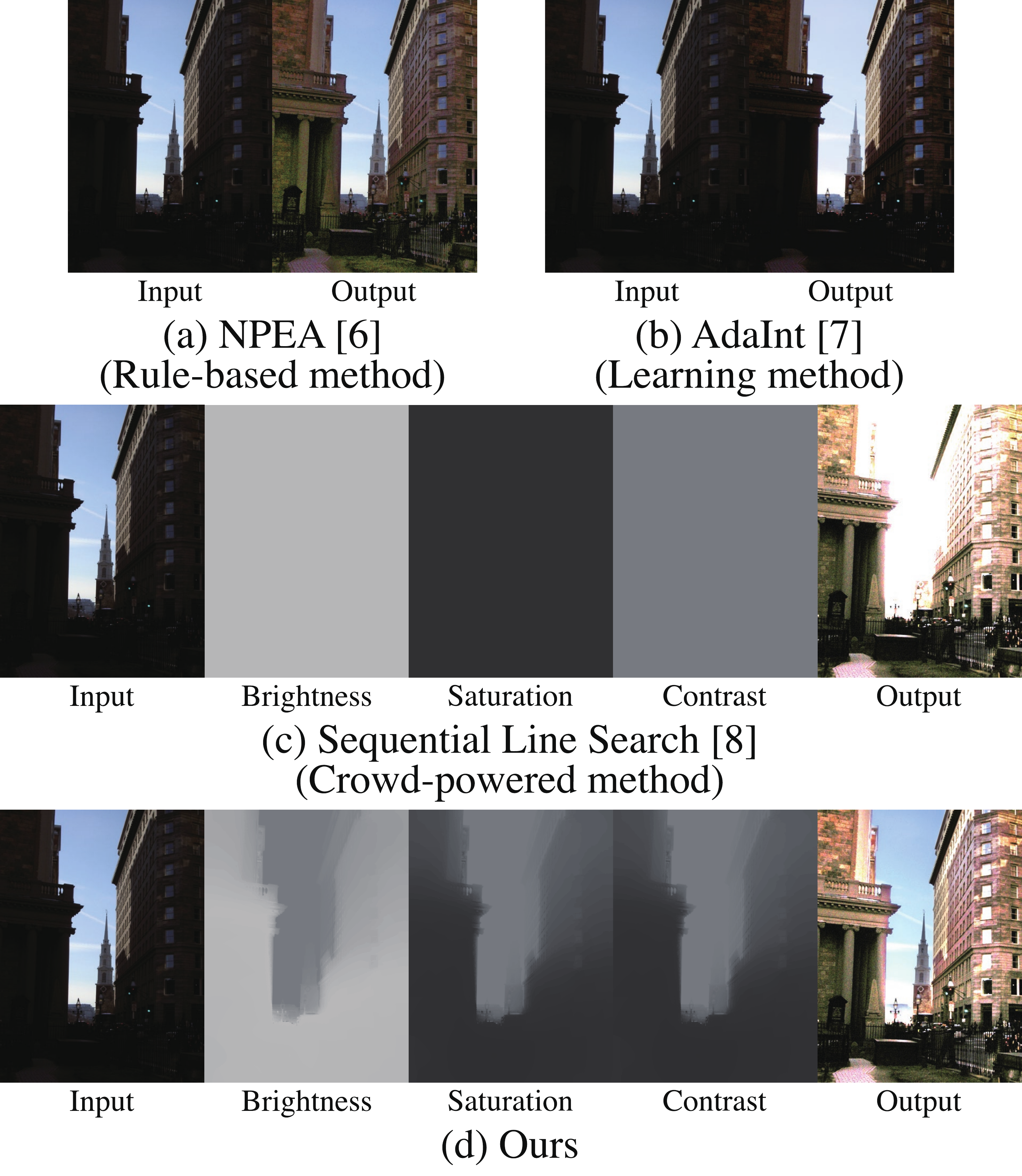}
  \caption{Comparisons with a previous rule-based method~\cite{wang2013naturalness},
  a previous learning method~\cite{yang2022adaint}, and a previous
  crowd-powered method~\cite{koyama2017sequential}.
  The three images in the middle in (c) and (d) show the parameters for the image editing functions.
  }
\label{fig1}
\end{figure}

Existing photo enhancement methods are classified into three categories:
rule-based methods, learning methods, and crowd-powered methods,
and we found that each method has limitations.
Rule-based methods decompose the input image into illumination and reflection.
By applying gamma correction or histogram equalization to the illumination,
the methods achieve local enhancement, which brightens only the dark areas.
However, in experiments using some rule-based methods~\cite{wang2013naturalness,fu2016fusion,fu2016weighted,guo2016lime,li2018structure},
we found that the performances are limited
because they can only apply the same effect to all input images
and cannot achieve content-aware enhancement.
In learning methods~\cite{ignatov2017dslr,wei2018deep,cai2018learning,wang2019underexposed,xu2020learning,afifi2021learning,yang2022adaint},
the enhancement model learns the translation
using numerous low- and high-quality images.
Although human-created paired datasets can be used to achieve content-aware enhancement,
locally-edited high-quality paired datasets do not exist because they are expensive to create.
Unpaired learning methods~\cite{chen2018deep,guo2020zero,jiang2021enlightengan} have been developed to reduce the cost of dataset creation;
however, they are inferior to paired learning methods.
In crowd-powered methods~\cite{koyama2017sequential,koyama2020sequential},
a task of adjusting multiple parameters for image editing functions is broken down into a sequence of
simple tasks, which is performed by crowd workers.
Content-aware enhancement is achieved by crowd workers,
but local enhancement is not possible because only globally consistent parameters are used.
In conclusion, the existing enhancement methods are either not content-aware or not local.

For content-aware local enhancement,
we propose a crowd-powered local enhancement method,
where we ask crowd workers to locally optimize the parameters for image editing functions.
To make it easier to locally optimize the parameters,
we use a novel active learning based local filter.
Only the parameters at a few key pixels need to be determined,
and the parameters at the other pixels are automatically predicted using a regression model.
To select the key pixels that are the most suitable for predicting the other parameters,
we propose to use active learning.
While general active learning is a technique to label numerous images efficiently,
we apply active learning to only one image considering the parameters of each pixel as the labeling target.
To achieve local photo enhancement using this local filter,
the parameters at the selected key pixels are optimized by crowd workers.
We break
down the task of optimizing multiple parameters into a sequence of single slider manipulation,
and crowd workers only need to adjust the single slider multiple times so that the image looks the best;
finally, the input image is locally enhanced considering the contents.
We show comparisons with a previous rule-based method~\cite{wang2013naturalness},
a previous learning method~\cite{yang2022adaint}, and a previous
crowd-powered method~\cite{koyama2017sequential}
in Figure~\ref{fig1}.
The rule-based method~\cite{wang2013naturalness} locally brightens the dark areas,
but the dark areas remain under-exposed;
the learning method~\cite{yang2022adaint} does not enhance the images locally.
While the previous crowd-powered method~\cite{koyama2017sequential} can only optimize the parameters globally,
our proposed method assigns different parameters to each pixel,
which enables higher quality enhancement.

To evaluate the proposed method, two experiments are designed.
First, to evaluate the active learning based local filter quantitatively,
we use an image quality assessment model to optimize the slider
and show that the filter achieves higher performance than existing local filters.
Second, we ask real crowd workers to adjust the slider
and show that the enhanced results obtain higher ratings than
the results by the existing enhancement methods in the user study.

This study makes the following contributions:
\begin{itemize}
\item To achieve content-aware local photo enhancement,
we propose a crowd-powered local enhancement method
that asks crowd workers to locally optimize the parameters for the image editing functions.
\item Our novel active learning based local filter makes it easier to locally optimize the parameters
by predicting the parameters from the pixels selected by active learning.
\item Our proposed filter outperforms the existing filters, and the
enhanced results obtain subjectively better evaluation than the results by the existing enhancement methods.
\end{itemize}

\begin{table}[!t]
  \caption{Pros and cons of each enhancement method.}
  \centering
  {\tabcolsep=0.88mm
  {\small
    \begin{tabular}{l|cccc}
      \hline
       & \begin{tabular}{c}Rule-based\\methods\end{tabular} & \begin{tabular}{c}Learning\\methods\end{tabular}& \begin{tabular}{c}Crowd-powered\\methods\end{tabular} &\rule{1mm}{0mm}Ours\rule{1mm}{0mm}\\
      \hline\hline
        \rule{0mm}{3.5mm}Content-aware & No & \underline{Yes} & \underline{Yes} & \underline{Yes}\\
        \rule{0mm}{3.5mm}Local & \underline{Yes} & No & No & \underline{Yes}\\
       \hline
    \end{tabular}
  }}
  \label{comparison_with_previous}
\end{table}

\section{Related Works}
Previous photo enhancement methods are classified into
rule-based methods, learning methods, and crowd-powered methods.
We summarize pros and cons in Table~\ref{comparison_with_previous} and
describe each method in the following sections.

\subsection{Rule-Based Methods}
Rule-based methods apply predefined effects to the input image.
Most of the recent methods are based on the Retinex theory~\cite{land1977retinex},
which decomposes the image into illumination (the brightness of the environment)
and reflection (the brightness change of the object surface).
Wang et al.~\cite{wang2013naturalness} proposed a bright-pass filter to
balance the detail and naturalness of the image.
Fu et al.~\cite{fu2016fusion} corrected the illumination map
by combining multiple illumination maps with different brightness levels.
Fu et al.~\cite{fu2016weighted} used a weighted variational model to simultaneously estimate illumination and reflection.
Guo et al.~\cite{guo2016lime} improved the consistency of illumination using a structure-aware smoothing model.
Li et al. ~\cite{li2018structure} added a noise map to improve the correction performance
of dark images containing intense noise.
These methods achieve local enhancement, which brightens only dark areas by
applying gamma correction or histogram equalization to the illumination.
However, the performance is limited because it does not apply different effects to each input image
considering the contents.

\subsection{Learning Methods\label{learning}}
In learning methods, the enhancement model learns the translation
using a large number of low- and high-quality images.
Most of the recent methods are based on convolutional neural networks (CNNs),
and Yan et al.~\cite{yan2016automatic} were the first to apply
CNNs to photo enhancement.
Wang et al.~\cite{wang2019underexposed} developed a novel loss function
that enables a spatially smooth enhancement.
Zhang et al.~\cite{zhang2019kindling} built a model inspired by rule-based methods.
Moran et al.~\cite{moran2020deeplpf} designed local parametric filters
for a lightweight model.
Kim et al.~\cite{kim2020global} combined global and local enhancement models.
He et al.~\cite{he2020conditional} reproduced image processing operations using
multilayer perceptrons.
Afifi et al.~\cite{afifi2021learning} proposed a coarse-to-fine framework
to enhance over- and under-exposed images.
Kim et al.~\cite{kim2021representative} developed representative color transform for details and high capacity.
Zhao et al.~\cite{zhao2021deep} adopted invertible neural networks for bidirectional feature learning.
Song et al.~\cite{song2021starenhancer} achieved style-aware enhancement.
Zheng et al.~\cite{zheng2021windowing} improved standard convolutions by integrating a local decomposition method.
Li et al.~\cite{li2021low} used a recursive unit to repeatedly unfold the input image for feature extraction.
Xu et al.~\cite{xu2022structure} presented a structure-texture aware network to fully consider the global structure and local detailed texture.
Dhara et al.~\cite{dhara2021exposedness} proposed a structure-aware exposedness estimation procedure for noise-suppressing enhancement.
For real-time photo enhancement,
Gharbi et al.~\cite{gharbi2017deep} applied a bilateral grid~\cite{chen2007real},
Lv et al.~\cite{lv2020fast} built light-weight CNNs,
Zeng et al.~\cite{zeng2020learning}, Wang et al.~\cite{wang2021real}, and Yang et al.~\cite{yang2022adaint}
used learnable 3D lookup tables,
and Zhang et al.~\cite{zhang2021star} used Transformer~\cite{vaswani2017attention}.

To train these models, paired datasets of low-quality original images and high-quality reference images
are necessary.
The MIT-Adobe 5K dataset~\cite{bychkovsky2011learning}
contains 5,000 pairs of original and expert-retouched images.
The DPED~\cite{ignatov2017dslr} consists of images taken by smartphone cameras and high-quality cameras.
The LOL dataset~\cite{wei2018deep} is composed of low- and normal-light image pairs.
The SID dataset~\cite{chen2018learning} is also composed of low- and normal-light pairs,
which are not sRGB images but raw sensor data.
To apply the SID dataset to sRGB images, the sRGB-SID dataset~\cite{xu2020learning} was created from the SID dataset.
Cai et al.~\cite{cai2018learning} generated
reference images by combining images with different exposure levels.
Wang et al.~\cite{wang2019underexposed} collected images under diverse lighting conditions that were retouched by experts.
However, these datasets have the following limitations.
The MIT-Adobe 5K dataset uses only a global filter for retouching,
and the reference images of the DPED, the LOL dataset, the SID dataset, and the sRGB-SID dataset
are created by changing the shooting condition globally,
which makes local enhancement impossible.
The quality of Cai et al.'s reference images is low
because blur and ghosting artifacts are contained.
Wang et al.'s dataset is not publicly available.

To reduce the cost of creating datasets,
methods which do not need paired images have been developed~\cite{deng2018aesthetic,chen2018deep,hu2018exposure,kosugi2020unpaired,guo2020zero,jiang2021enlightengan,zhang2021rellie,zhao2021retinexdip,liang2022self}.
Particularly, Guo et al.~\cite{guo2020zero} formulated non-reference loss functions.
Jiang et al.~\cite{jiang2021enlightengan} introduced a local discriminator into
a generative adversarial network and trained the network using unpaired images.
These models achieved local enhancement,
but their performance is inferior to that of paired methods.

\subsection{Crowd-powered Methods}
In crowd-powered methods,
the parameters for the image editing functions are determined by crowd workers to obtain enhanced results.
Because the multiple image editing functions are intricately interrelated,
it is difficult for non-expert crowd workers to adjust the multiple parameters simultaneously.
To solve this problem, Sequential Line Search (SLS)~\cite{koyama2017sequential}
breaks down the optimization problem of the multiple parameters into a sequence of single-slider adjustments,
which can be easily performed by crowd workers.
Sequential Gallery~\cite{koyama2020sequential}
replaces the single slider with a two-dimensional search space to make the optimization process more efficient,
but the performance is evaluated in simulated experiments because the task is complex for crowd workers.
Content-aware enhancement is achieved by crowd workers,
but local enhancement is not possible because only a global filter is used.

\begin{figure*}[t]
  \centering
  \includegraphics[width=1\hsize]{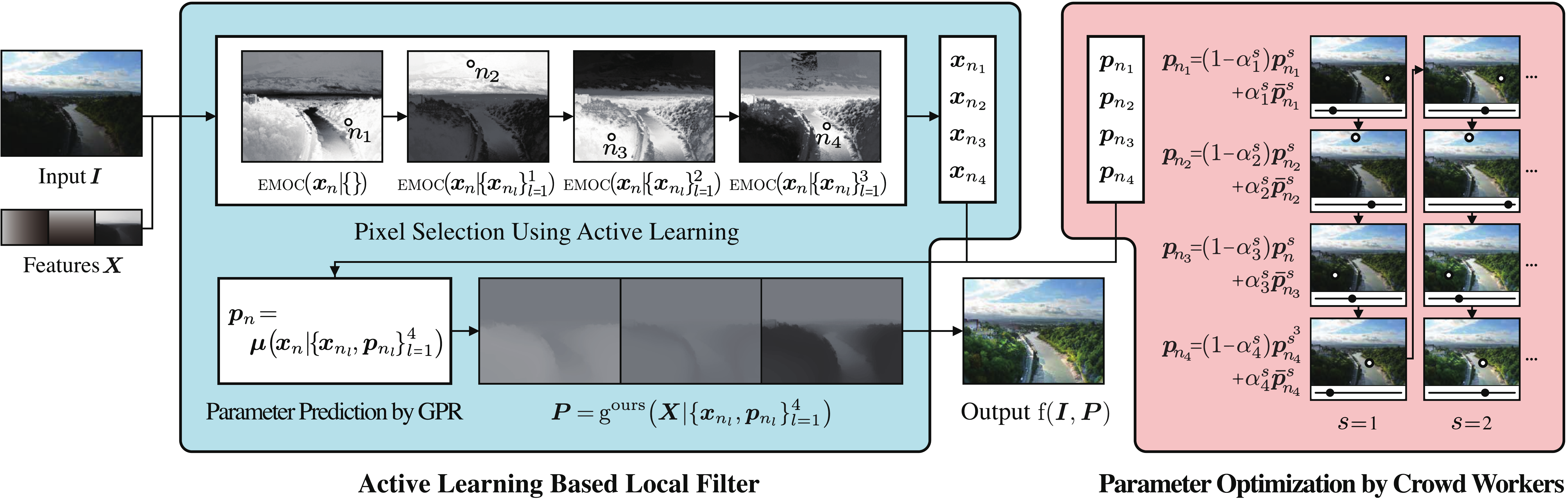}
\caption{Overview of our method, where $L=4$.
In our active learning based local filter,
$L$ key pixels $n_1,...,n_L$ are selected;
only parameters $\bm{p}_{n_1},...,\bm{p}_{n_L}$ need to be determined,
and the other parameters are automatically predicted by a regression model.
To make the filter highly expressive, we define the pixel features ${\bm X}$
as a combination of the xy-coordinates and an illumination map ${\bm t}$,
and to make the filter easy to optimize,
the key pixels $n_1,...,n_L$ are selected using an active learning method, EMOC~\cite{kading2018active}.
The parameters $\bm{p}_{n_1},...,\bm{p}_{n_L}$ are optimized independently by crowd workers using SLS~\cite{koyama2017sequential};
consequently, we can obtain the best enhanced results.}
\label{proposedmethod}
\end{figure*}

\begin{figure*}[t]
  \centering
  \includegraphics[width=1\hsize]{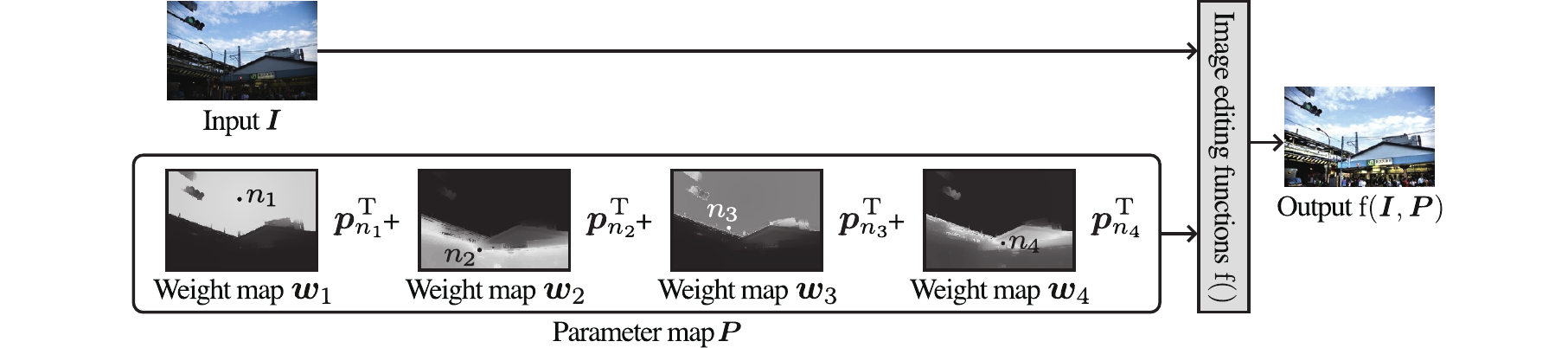}
\caption{Visualization of our active learning based local filter
where the number of the key pixels $L=4$.
$n_1,...,n_4$ are the key pixels selected using active learning,
and $\bm{p}_{n_1},...,\bm{p}_{n_4}$ are parameters at the key pixels.
The parameter map ${\bm P}$ is represented as ${\bm P}={\bm w}_1\bm{p}_{n_1}^{\rm T} + \cdots +{\bm w}_4\bm{p}_{n_4}^{\rm T}$, and the edited result is obtained as ${\rm f}({\bm I}, {\bm P})$.
\label{our_filter}}
\end{figure*}

\section{Proposed Method}
Our goal is to enhance the aesthetic quality of an input
image by changing the color and brightness.
Because the existing photo enhancement methods cannot consider the contents or
cannot enhance images locally,
we propose a crowd-powered local enhancement method
for content-aware local enhancement.
We denote the input image as ${\bm I} = [{\bm i_1} ... {\bm i_N}]^{\rm T}\in\mathbb{R}^{N\times3}$
and a parameter map for the image editing functions as ${\bm P} = [{\bm p_1} ... {\bm p_N}]^{\rm T}\in\mathbb{R}^{N\times M}$,
where $N$ is the number of pixels in the input image,
and $M$ is the number of the image editing functions,
which change the color properties such as brightness and contrast.
The edited result is denoted as ${\rm f}({\bm I}, {\bm P})$, where
${\rm f}()$ applies the effects by $M$ types of image editing functions
to ${\bm I}$ based on ${\bm P}$.
The goal of our method is to maximize the aesthetic quality of ${\rm f}({\bm I}, {\bm P})$
by optimizing ${\bm P}$.

The previous crowd-powered method, SLS~\cite{koyama2017sequential},
uses ${\rm f()}$ as a global filter,
\begin{equation}\label{tile_}
	{\rm f}({\bm I}, {\bm P}) = {\rm f}\Bigl({\bm I}, {\rm g^{global}}\bigl(\bm{p}^{\rm global}\bigr)\Bigr),
\end{equation}
where $\bm{p}^{\rm global}\in\mathbb{R}^{M}$,
and the function ${\rm g^{global}()}$ assigns the same parameters to all the pixels,
\begin{equation}\label{tile}
  {\bm P} =  {\rm g^{global}}\bigl(\bm{p}^{\rm global}\bigr) = {\bm 1}(\bm{p}^{\rm global})^{\rm T},
\end{equation}
where ${\bm 1\in\mathbb{R}^N}$ is an all-ones vector.
To enhance the input image,
they ask crowd workers to optimize $\bm{p}^{\rm global}$.
Because the $M$ image editing functions in ${\rm f()}$ are intricately related to each other,
it is difficult for non-expert crowd workers to optimize each element $p^{\rm global}_1, ..., p^{\rm global}_M$
simultaneously.
To solve this problem, SLS breaks down the task of optimizing multiple parameters
into a sequence of single slider manipulation.
SLS first generates random parameters $\bm{p}^{\rm 1}$ and $\bm{\bar{p}}^{\rm 1}$,
and the crowd workers adjust a single slider that controls $\alpha_1\in[0,1]$
to make the following parameter map ${\bm P}$ the best,
\begin{equation}\label{aa}
	{\bm P} = {\rm g^{global}}\bigl((1-\alpha_1)\bm{p}^{\rm 1} + \alpha_1\bm{\bar{p}}^{\rm 1}\bigr).
\end{equation}
The best $\alpha_1$ is defined as $\alpha_1^*$, and the best parameter
$(1-\alpha^*_1)\bm{p}^{\rm 1} + \alpha^*_1\bm{\bar{p}}^{\rm 1}$ is defined as $\bm{p}^{\rm 2}$.
SLS presents the most informative parameter $\bm{\bar{p}}^{\rm 2}$
based on the previous parameters $\bigl\{\bm{p}^{\rm 1},\bm{\bar{p}}^{\rm 1},$
$\bm{p}^{\rm 2}\bigr\}$,
\begin{equation}
\bm{\bar{p}}^{\rm 2} = {\rm SLS}\bigl(\bigl\{\bm{p}^{\rm 1},\bm{\bar{p}}^{\rm 1},\bm{p}^{\rm 2}\bigr\}\bigr).
\end{equation}
The crowd workers choose the best parameter $\bm{p}^{\rm 3}$
from the linear interpolation of $\bm{p}^{\rm 2}$
and $\bm{\bar{p}}^{\rm 2}$.
To sum up, the following processes are repeated from $s=1$ to $s=S$:
\begin{enumerate}
  \setlength{\itemsep}{0.2cm}
   \item Crowd workers adjust $\alpha_s\in[0,1]$ to make the following parameter map ${\bm P}$ the best,
   \begin{equation}
   	{\bm P} = {\rm g^{global}}\bigl((1-\alpha_s)\bm{p}^{\rm s} + \alpha_s\bm{\bar{p}}^{\rm s}\bigr).
   \end{equation}
   The best $\alpha_s$ is denoted as $\alpha_s^*$.
   \item $\bm{p}^{\rm s+1} = (1-\alpha^*_s)\bm{p}^{\rm s} + \alpha^*_s\bm{\bar{p}}^{\rm s}$.
   \item  $\bm{\bar{p}}^{\rm s+1} = {\rm SLS}\bigl(\bigl\{\bm{p}^{\rm 1},\bm{\bar{p}}^{\rm 1},...,\bm{p}^{\rm s},\bm{\bar{p}}^{\rm s},\bm{p}^{\rm s+1}\bigr\}\bigr)$.
\vspace{0.1cm}
\end{enumerate}
Finally, optimized $\bm{p}^{\rm global}$ is obtained as $\bm{p}^{S+1}$.

To optimize ${\bm P}$ with single slider manipulation
allowing ${\bm p_n}$ to take different values for each pixel,
we propose a novel local filter.
There are two important aspects in designing the local filter:
the filter should be (i) {\bf highly expressive} and (ii) {\bf easy to optimize}.
For example, the global filter is easy to optimize but not expressive,
and the pixel-wise optimization of the parameters provides a high expressive power but
needs a large optimization cost.
Because crowdsourcing is time-consuming and expensive,
the optimization should be completed in as few iterations as possible.
To satisfy the two aspects,
we propose an active learning based local filter.
We select $L$ key pixels $n_1,...,n_L$ from $\bm I$.
Only parameters $\bm{p}_{n_1},...,\bm{p}_{n_L}$ need to be determined,
and the other parameters are automatically predicted by a regression model.
We denote the pixel features as ${\bm X} = [{\bm x_1} ... {\bm x_N}]^{\rm T}$, and our filter is represented as
\begin{equation}
{\rm f}({\bm I}, {\bm P}) = {\rm f}\Bigl({\bm I}, {\rm g^{ours}}\bigl({\bm X} |\{\bm{x}_{n_l}, \bm{p}_{n_l}\}^L_{l=1}\bigr)\Bigr).
\end{equation}
In ${\rm g^{ours}}$, $\bm{p}_{n}$ is predicted as
\begin{equation}\label{mu}
\bm{p}_{n} = {\bm \mu}\bigl(\bm{x}_{n}|\{\bm{x}_{n_l}, \bm{p}_{n_l}\}^L_{l=1}\bigr),
\end{equation}
where ${\bm \mu}()$ is the regression model.
The design of $\bm{x}_{n}$ makes it possible to achieve a highly expressive filter
that depends on the spatial location and brightness of the area.
To make it as easy as possible to optimize the filter,
key pixels $n_1,...,n_L$ are selected using active learning.
Active learning is originally used to label a large number of images efficiently; however,
in our method, we apply active learning to only one image
to select the key pixels which are the most suitable for predicting the other parameters.
$\bm{p}_{n_1},...,\bm{p}_{n_L}$ are optimized independently using SLS;
consequently, we can obtain the best enhanced results.
The overview of our method is shown in Figure~\ref{proposedmethod},
and we present a detailed explanation in the following sections.

\subsection{Active Learning Based Local Filter}
\subsubsection{Parameter Prediction by Gaussian Process Regression}
We train a regression model using
the features $\bm{x}_{n_1},...,\bm{x}_{n_L}$ and the parameters
$\bm{p}_{n_1},...,\bm{p}_{n_L}$,
and the model predicts $\bm{p}_{n}$ from $\bm{x}_{n}$
as in Eq.~(\ref{mu}).
As a regression model, we use Gaussian Process Regression (GPR)~\cite{rasmussen2003gaussian},
which is generally used for active learning.
Using GPR, $\bm{p}_{n}$ is represented as follows,
\begin{equation}\label{mu_}
	\bm{p}_{n} = {\bm \mu}\bigl(\bm{x}_{n}|\{\bm{x}_{n_l}, \bm{p}_{n_l}\}^L_{l=1}\bigr)
	= \bigl({\bm k}_{n}^{\rm T}{\bm K}^{-1}{\bm Q}\bigr)^{\rm T},
\end{equation}
where
\begin{equation}
\bm{k}_n = \begin{bmatrix}
\kappa({\bm x}_{n}, {\bm x}_{n_1}) & \cdots & \kappa({\bm x}_{n}, {\bm x}_{n_L})
\end{bmatrix}^{\rm T},
\end{equation}
\begin{equation}
{\bm K} = \begin{bmatrix}
\kappa({\bm x}_{n_1}, {\bm x}_{n_1})+r&\cdots&\kappa({\bm x}_{n_1}, {\bm x}_{n_L})\\
\vdots&\ddots&\vdots\\
\kappa({\bm x}_{n_L}, {\bm x}_{n_1})&\cdots&\kappa({\bm x}_{n_L}, {\bm x}_{n_L})+r \\
\end{bmatrix},
\end{equation}
\begin{equation}
	{\bm Q} = \begin{bmatrix}
	\bm{p}_{n_1}&\cdots&\bm{p}_{n_L}
\end{bmatrix}^{\rm T},
\end{equation}
$\kappa()$ is a kernel function, and $r$ is a regularization term.

Eq.~(\ref{mu_}) seems complex, but our filter can be represented simply.
We denote ${\bm W}=[({\bm k}_1^{\rm T}{\bm K}^{-1})^{\rm T}...({\bm k}_N^{\rm T}{\bm K}^{-1})^{\rm T}]^{\rm T}\in\mathbb{R}^{N\times L}$,
and based on Eq.~(\ref{mu_}), the parameter map ${\bm P}$ is represented as
\begin{equation}\begin{split}
	{\bm P} &= [{\bm p_1} ... {\bm p_N}]^{\rm T} \\
          &= [\bigl({\bm k}_{1}^{\rm T}{\bm K}^{-1}{\bm Q}\bigr)^{\rm T} ... \bigl({\bm k}_{N}^{\rm T}{\bm K}^{-1}{\bm Q}\bigr)^{\rm T}]^{\rm T}\\
          &=[({\bm k}_1^{\rm T}{\bm K}^{-1})^{\rm T}...({\bm k}_N^{\rm T}{\bm K}^{-1})^{\rm T}]^{\rm T}{\bm Q} \\
          &={\bm W}{\bm Q} \\
          &={\bm w}_1\bm{p}_{n_1}^{\rm T} + \cdots +{\bm w}_L\bm{p}_{n_L}^{\rm T},
\end{split}\end{equation}
where ${\bm w}_1,...,{\bm w}_L$ are column vectors of ${\bm W}$, {\it i.e.},
${\bm W}=[{\bm w}_1...{\bm w}_L]$.
In this representation, ${\bm w}_1,...,{\bm w}_L$ can be regarded as weight maps for $\bm{p}_{n_1},...,\bm{p}_{n_L}$,
and the parameter map ${\bm P}$ can be calculated as a weighted sum of $\bm{p}_{n_1},...,\bm{p}_{n_L}$.
We show the visualization of our filter in Figure~\ref{our_filter}.
The weight maps ${\bm w}_1,...,{\bm w}_L$ have different shapes of local weights.
By using GPR, we can obtain these local weight maps.

\begin{figure}[t]
  \centering
  \includegraphics[width=1\hsize]{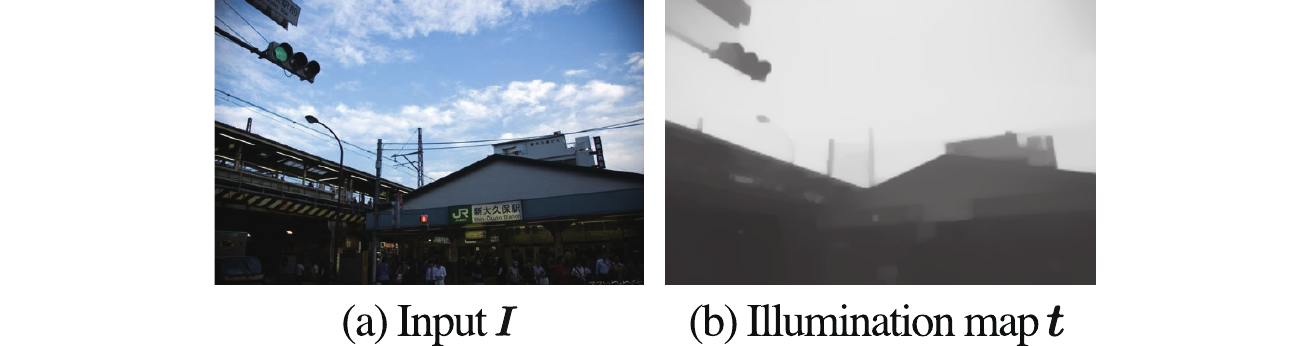}
\caption{An example of the illumination map generated by LIME~\cite{guo2016lime}.
\label{illumination_map}}
\end{figure}

\begin{figure}[t]
  \centering
  \includegraphics[width=1\hsize]{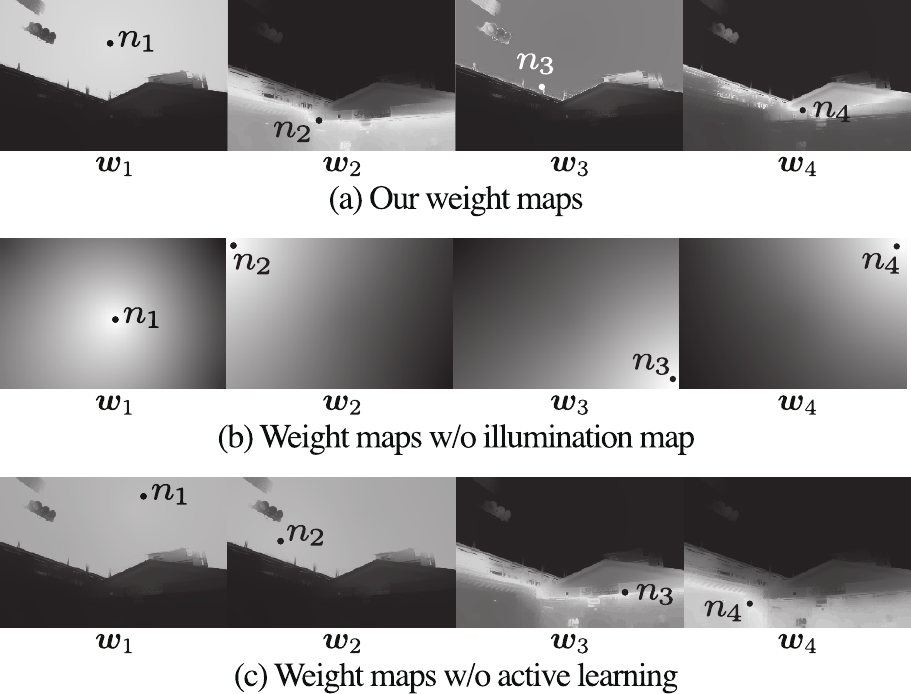}
\caption{Visualization of weight maps where the number of the key pixels $L=4$. $n_1,...,n_4$ are the key pixels.
\label{weight_maps}}
\end{figure}

\subsubsection{Design of the Pixel Features}
To achieve a high expressive power, the design of the pixel feature $\bm{x}_{n}$ is important.
If only the xy-coordinates of $n$ are used as the feature $\bm{x}_{n}$,
GPR cannot achieve edge-aware prediction.
To achieve edge-aware prediction, we propose the use of an illumination map
${\bm t} = [t_1...t_N]^{\rm T}$ of ${\bm I}$,
which is an estimation of the brightness of the environment,
and we use LIME~\cite{guo2016lime} as the illumination estimation method.
An example of the illumination map is shown in Figure~\ref{illumination_map}.
We define $\bm{x}_{n}$ as a combination of the xy-coordinates and $t_n$.
We show the weight maps with and without the illumination map in Figures~\ref{weight_maps}(a) and (b), respectively.
When the illumination map is not used,
only gradation-like weight maps are obtained.
Because GPR can achieve edge-aware prediction by using the illumination map,
we can obtain edge-aware weight maps.

\subsubsection{Pixel Selection Using Active Learning}
To select the key pixels ${n_1},...,{n_L}$, which are the most suitable for the prediction,
we propose the application of an active learning method.
After selecting $L'$ pixels ${n_1},...,{n_{L'}}$,
we can find the next pixel to be selected, $n_{L'+1}$, using the active learning method.
Among existing active learning methods for regression~\cite{freund1997selective,sugiyama2006active,yu2010passive,hu2010egal,cai2013maximizing,goetz2018active,tsymbalov2018dropout,wu2018pool,kading2018active,wu2019active,zhang2020graph},
we can use only label-independent methods,
because the labels $\bm{p}_{n_1},...,\bm{p}_{n_L}$ are changed during the optimization process.
We experimentally find that Kading et al.'s~\cite{kading2018active} method
outperforms other available methods;
therefore, we use it in our method.
They define the Expected Model Output Changes (EMOC) as an evaluation metric for the next sample to be selected.
EMOC is the expected value of the GPR model's output changes
with and without a novel pixel,
and we can select the best pixel for prediction by selecting the pixel whose EMOC is the highest.
The calculation of EMOC is presented in the original paper,
but it is important that the EMOC does not depend on the parameters ${\bm p}_{n_1},...,{\bm p}_{n_{L'}}$.
Therefore, the key pixels can be selected before the parameters are determined.
The next pixel to choose $\bm{x}_{n_{L'+1}}$ is calculated as follows,
\begin{equation}
	n_{L'+1} = {\rm argmax}_{n}~{\rm EMOC}\bigl(\bm{x}_{n}|\{\bm{x}_{n_l}\}^{L'}_{l=1}\bigr).
\end{equation}
By repeating this calculation from $L'=0$ to $L-1$,
the key pixels $n_1,...,n_L$ are obtained.
Note that no human annotation is required for the selection of the key pixels ${n_1},...,{n_L}$;
EMOC is automatically calculated based on the features of an input image.

We show the weight maps where the key pixels are selected with and without the active learning
in Figures~\ref{weight_maps}(a) and (c), respectively.
When the key pixels are selected without the active learning ({\it i.e.}, randomly), ${\bm w}_1$ and ${\bm w}_2$ have the similar shapes,
which are inefficient.
By using the EMOC, we can obtain different shapes of weight maps.

\begin{figure*}[t]
 \centering
 \includegraphics[width=1\hsize]{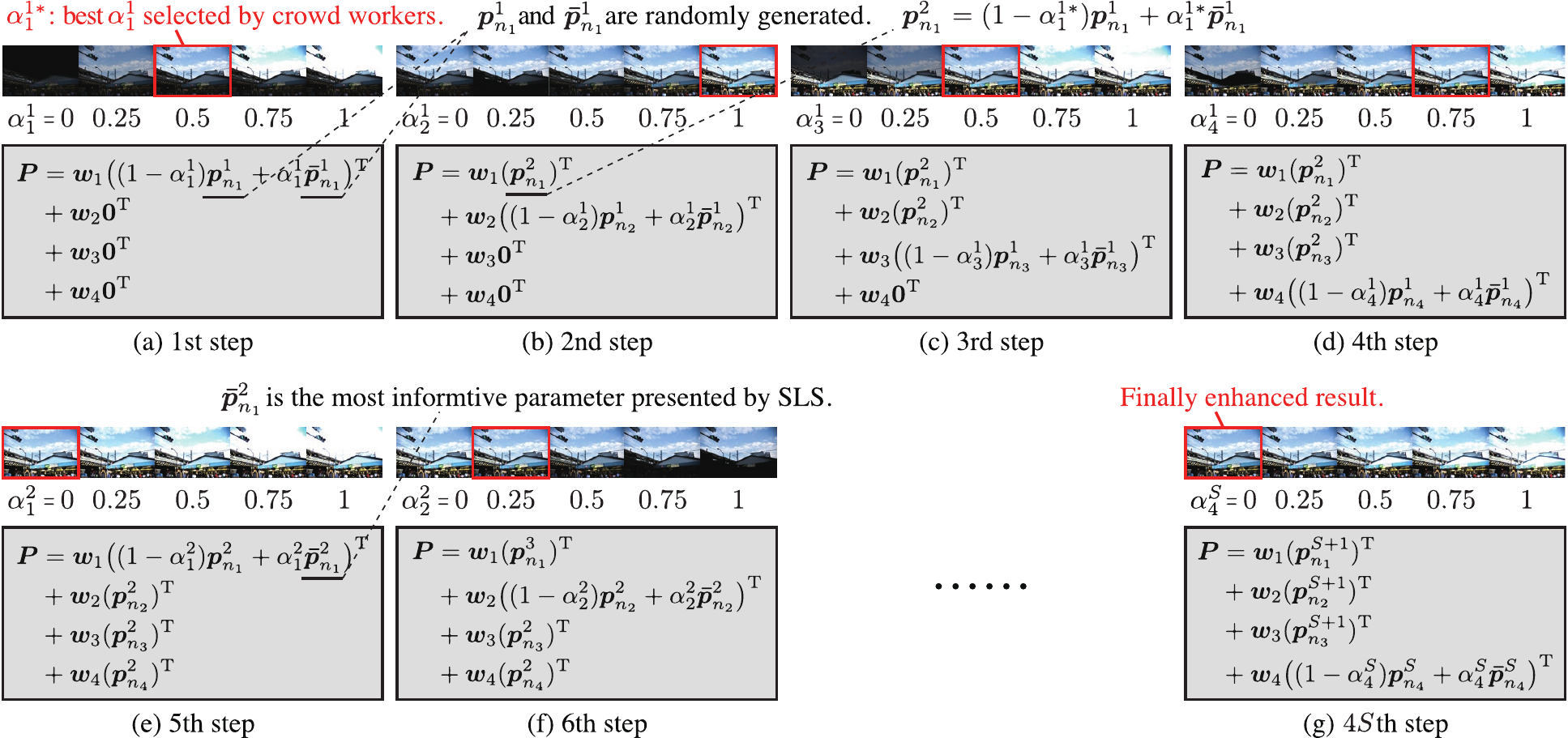}
\caption{
Visualization of the optimization process of the parameter map ${\bm P}$
where the number of the key pixels $L=4$.
In the first step (a), $\bm{p}^1_{n_1}$ and $\bm{\bar p}^{1}_{n_1}$ are randomly generated,
and crowd workers adjust $\alpha^1_1$ to make the parameter map ${\bm P} = {\bm w}_1\bigl((1 - \alpha^1_1)\bm{p}^1_{n_1} + \alpha^1_1\bm{\bar p}^{1}_{n_1}\bigr)^{\rm T}+{\bm w}_2\bm{0}^{\rm T}+{\bm w}_3\bm{0}^{\rm T}+{\bm w}_4\bm{0}^{\rm T}$
best.
The adjusted $\alpha^1_1$ is denoted as $\alpha^{1*}_1$, and the adjusted parameter is denoted as
$\bm{p}^{2}_{n_1} = (1 - \alpha^{1*}_1)\bm{p}^1_{n_1}  + \alpha^{1*}_1\bm{\bar p}^{1}_{n_1}$.
In the second step (b), crowd workers adjust $\alpha^1_2$ to make the parameter map
${\bm P} = {\bm w}_1(\bm{p}^{2}_{n_1})^{\rm T}+{\bm w}_2\bigl((1 - \alpha^1_2)\bm{p}^1_{n_2} + \alpha^1_2\bm{\bar p}^{1}_{n_2}\bigr)^{\rm T}+{\bm w}_3\bm{0}^{\rm T}+{\bm w}_4\bm{0}^{\rm T}$ best.
In the same way, crowd workers adjust $\alpha^1_3$ and $\alpha^1_4$ in the third and fourth steps.
In the fifth step (e), the most informative parameter $\bm{\bar p}^{2}_{n_1}$ is obtained as
$\bm{\bar p}^{2}_{n_1} = {\rm SLS}\bigl(\bigl\{\bm{p}^1_{n_1},\bm{\bar p}^{1}_{n_1},\bm{p}^2_{n_1}\bigr\}\bigr)$,
and crowd workers adjust $\alpha^2_1$ to make the parameter map
${\bm P} = {\bm w}_1\bigl((1 - \alpha^2_1)\bm{p}^2_{n_1} + \alpha^2_1\bm{\bar p}^2_{n_1}\bigr)^{\rm T} + {\bm w}_2(\bm{p}^{2}_{n_2})^{\rm T}+{\bm w}_3(\bm{p}^{2}_{n_3})^{\rm T}+{\bm w}_4(\bm{p}^{2}_{n_4})^{\rm T}$ best.
Finally, after the $4S$th step, the enhanced result is obtained as ${\bm P} = {\bm w}_1(\bm{p}_{n_1}^{S+1})^{\rm T} + {\bm w}_2(\bm{p}_{n_2}^{S+1})^{\rm T}+{\bm w}_3(\bm{p}_{n_3}^{S+1})^{\rm T} +{\bm w}_4(\bm{p}_{n_4}^{S+1})^{\rm T}$.
Although the images are in one row for visualization,
crowd workers can change the image using a slider as shown in Figure~\ref{interface}(b).
}
\label{slider}
\end{figure*}

\subsection{Parameter Optimization by Crowd Workers}
The goal of our method is to maximize the aesthetic quality of
${\rm f}({\bm I}, {\bm P}) = {\rm f}({\bm I}, {\bm w}_1\bm{p}_{n_1}^{\rm T} + \cdots +{\bm w}_L\bm{p}_{n_L}^{\rm T})$.
For this purpose, we apply SLS~\cite{koyama2017sequential} to each parameter $\bm{p}_{n_l}$ independently.
We first generate two random parameters $\bm{p}^1_{n_1}$ and $\bm{\bar p}^{1}_{n_1}$,
and the crowd workers adjust $\alpha^1_1$ by controlling a slider to make the following parameter map best,
\begin{equation}\label{slider1_1}
  {\bm P} = {\bm w}_1\bigl((1 - \alpha^1_1)\bm{p}^1_{n_1} + \alpha^1_1\bm{\bar p}^{1}_{n_1}\bigr)^{\rm T}+{\bm w}_2\bm{0}^{\rm T}+\cdots+{\bm w}_L\bm{0}^{\rm T},
\end{equation}
where $\bm{0}$ is a zero vector.
In Eq.~(\ref{slider1_1}), $\bm{p}_{n_1}$ is optimized while $\bm{p}_{n_2}, ..., \bm{p}_{n_L}$ are fixed to $\bm{0}$.
The parameter map ${\bm P}$ is updated when changing $\alpha^1_1$,
and the crowd workers adjust $\alpha^1_1$ to make the enhanced result ${\rm f}({\bm I}, {\bm P})$ best.
We define the optimized $\alpha^1_1$ as $\alpha^{1*}_1$, and the optimized parameter $\bm{p}^{2}_{n_1}$ is
represented as
\begin{equation}\label{aaa}
  \bm{p}^{2}_{n_1} = (1 - \alpha^{1*}_1)\bm{p}^1_{n_1}  + \alpha^{1*}_1\bm{\bar p}^{1}_{n_1}.
\end{equation}
Subsequently, we generate two random parameters $\bm{p}^1_{n_2}$ and $\bm{\bar p}^{1}_{n_2}$,
and the crowd workers adjust $\alpha^1_2$ by controlling a slider to make the following parameter map best,
\begin{equation}\begin{split}
    {\bm P} = {\bm w}_1(\bm{p}^{2}_{n_1})^{\rm T} &+ {\bm w}_2\bigl((1 - \alpha^1_2)\bm{p}^1_{n_2} + \alpha^1_2\bm{\bar p}^{1}_{n_2}\bigr)^{\rm T}\\
    &+{\bm w}_3\bm{0}^{\rm T}+\cdots+{\bm w}_L\bm{0}^{\rm T},
\end{split}\end{equation}
where $\bm{p}_{n_2}$ is optimized while $\bm{p}_{n_1}$ is fixed to $\bm{p}^{2}_{n_1}$,
and $\bm{p}_{n_3}, ..., \bm{p}_{n_L}$ are fixed to $\bm{0}$.
In the same way, the crowd workers adjust $\alpha^1_l$ by controlling a slider to make the following parameter map best,
\begin{equation}\label{slider1}\begin{split}
    {\bm P} = &{\bm w}_1(\bm{p}^{2}_{n_1})^{\rm T} + \cdots+ {\bm w}_{l-1}(\bm{p}^{2}_{n_{l-1}})^{\rm T}\\
    &+{\bm w}_l\bigl((1 - \alpha^1_l)\bm{p}^1_{n_l} + \alpha^1_l\bm{\bar p}^{1}_{n_l}\bigr)^{\rm T}\\
    &+{\bm w}_{l+1}\bm{0}^{\rm T} + \cdots+ {\bm w}_L\bm{0}^{\rm T}.\\
\end{split}\end{equation}
This process is repeated from $l=1$ to $L$.
When this process for $l=L$ is finished,
the most informative parameter $\bm{\bar p}^{2}_{n_l}$ can be obtained using SLS
based on $\bigl\{\bm{p}^1_{n_l},\bm{\bar p}^{1}_{n_l},\bm{p}^2_{n_l}\bigr\}$,
\begin{equation}
\bm{\bar p}^{2}_{n_l} = {\rm SLS}\bigl(\bigl\{\bm{p}^1_{n_l},\bm{\bar p}^{1}_{n_l},\bm{p}^2_{n_l}\bigr\}\bigr).
\end{equation}
The crowd workers then select the best parameter
from the linear interpolation of $\bm{p}^2_{n_l}$ and $\bm{\bar p}^{2}_{n_l}$ repeatedly.
For each $\bm{p}_{n_l}$, the slider manipulation is repeated $S$ times.
When adjusting a slider for $\bm{p}_{n_l}$ for the $s$th time ($s\ge2$), the crowd workers adjust $\alpha^s_l\in[0,1]$ to make the following parameter map best,
\begin{equation}\label{slider2}\begin{split}
  {\bm P} = &{\bm w}_1(\bm{p}^{s+1}_{n_1})^{\rm T} + \cdots+ {\bm w}_{l-1}(\bm{p}^{s+1}_{n_{l-1}})^{\rm T}\\
  &+{\bm w}_l\bigl((1 - \alpha^s_l)\bm{p}^s_{n_l} + \alpha^s_l\bm{\bar p}^{s}_{n_l}\bigr)^{\rm T}\\
  &+{\bm w}_{l+1}(\bm{p}^{s}_{n_{l+1}})^{\rm T} + \cdots+ {\bm w}_L(\bm{p}^{s}_{n_{L}})^{\rm T}.\\
\end{split}\end{equation}
Finally, the optimized parameter map ${\bm P}$ is obtained as
\begin{equation}
  {\bm P} = {\bm w}_1(\bm{p}_{n_1}^{S+1})^{\rm T} + \cdots +{\bm w}_L(\bm{p}_{n_L}^{S+1})^{\rm T}.
\end{equation}
We show the visualization of the optimization process in Figure~\ref{slider}
and the pseudocode of our method in Algorithm~\ref{alg1}.

\renewcommand{\algorithmicrequire}{\textbf{Input:}}
\renewcommand{\algorithmicensure}{\textbf{Output:}}
\begin{figure}[!t]
\begin{algorithm}[H]
    \caption{Proposed method}
    \label{alg1}
    \begin{algorithmic}[1]
    \REQUIRE Image ${\bm I} = [{\bm i_1} ... {\bm i_N}]^{\rm T}$, Number of key pixels $L$, Number of slider manipulation $S$
    \ENSURE Optimized parameter map ${\bm P} = [{\bm p_1} ... {\bm p_N}]^{\rm T}$
    \STATE An illumination map ${\bm t} = [t_1...t_N]^{\rm T}$ is generated by \cite{guo2016lime}.
    \STATE Pixel feature $\bm{x}_{n}$ is defined as a combination of the xy-coordinates and $t_n$.
    \FOR {$L'=0$ to $L-1$}
      \STATE $n_{L'+1} \leftarrow {\rm argmax}_{n}~{\rm EMOC}\bigl(\bm{x}_{n}|\{\bm{x}_{n_l}\}^{L'}_{l=1}\bigr)$
    \ENDFOR
    \STATE ${\bm W}=[{\bm w}_1...{\bm w}_L]\leftarrow[({\bm k}_1^{\rm T}{\bm K}^{-1})^{\rm T}...({\bm k}_N^{\rm T}{\bm K}^{-1})^{\rm T}]^{\rm T}$
    \FOR {$s=1$ to $S$}
      \FOR {$l=1$ to $L$}
        \IF{$s = 1$}
          \STATE Random parameters $\bm{p}^1_{n_l}$ and $\bm{\bar p}^{1}_{n_l}$ are generated.
          \STATE Crowd workers adjust $\alpha^1_l\in[0,1]$ to make Eq.~(\ref{slider1}) best,
          and the best $\alpha^1_l$ is denoted as $\alpha^{1*}_l$.
        \ELSE
          \STATE Crowd workers adjust $\alpha^s_l\in[0,1]$ to make Eq.~(\ref{slider2}) best,
          and the best $\alpha^s_l$ is denoted as $\alpha^{s*}_l$.
        \ENDIF
        \STATE $\bm{p}^{s+1}_{n_l} \leftarrow  (1 - \alpha^{s*}_l)\bm{p}^s_{n_l} + \alpha^{s*}_l\bm{\bar p}^{s}_{n_l}$
        \STATE $\bm{\bar p}^{s+1}_{n_l} \leftarrow {\rm SLS}\bigl(\bigl\{\bm{p}^1_{n_l},\bm{\bar p}^{1}_{n_l},...,\bm{p}^{s}_{n_l},\bm{\bar p}^{s}_{n_l},\bm{p}^{s+1}_{n_l}\bigr\}\bigr)$
      \ENDFOR
    \ENDFOR
    \STATE ${\bm P} \leftarrow {\bm w}_1(\bm{p}_{n_1}^{S+1})^{\rm T} + \cdots +{\bm w}_L(\bm{p}_{n_L}^{S+1})^{\rm T}$
    \end{algorithmic}
\end{algorithm}
\end{figure}

\begin{figure}[t]
 \centering
 \includegraphics[width=1\hsize]{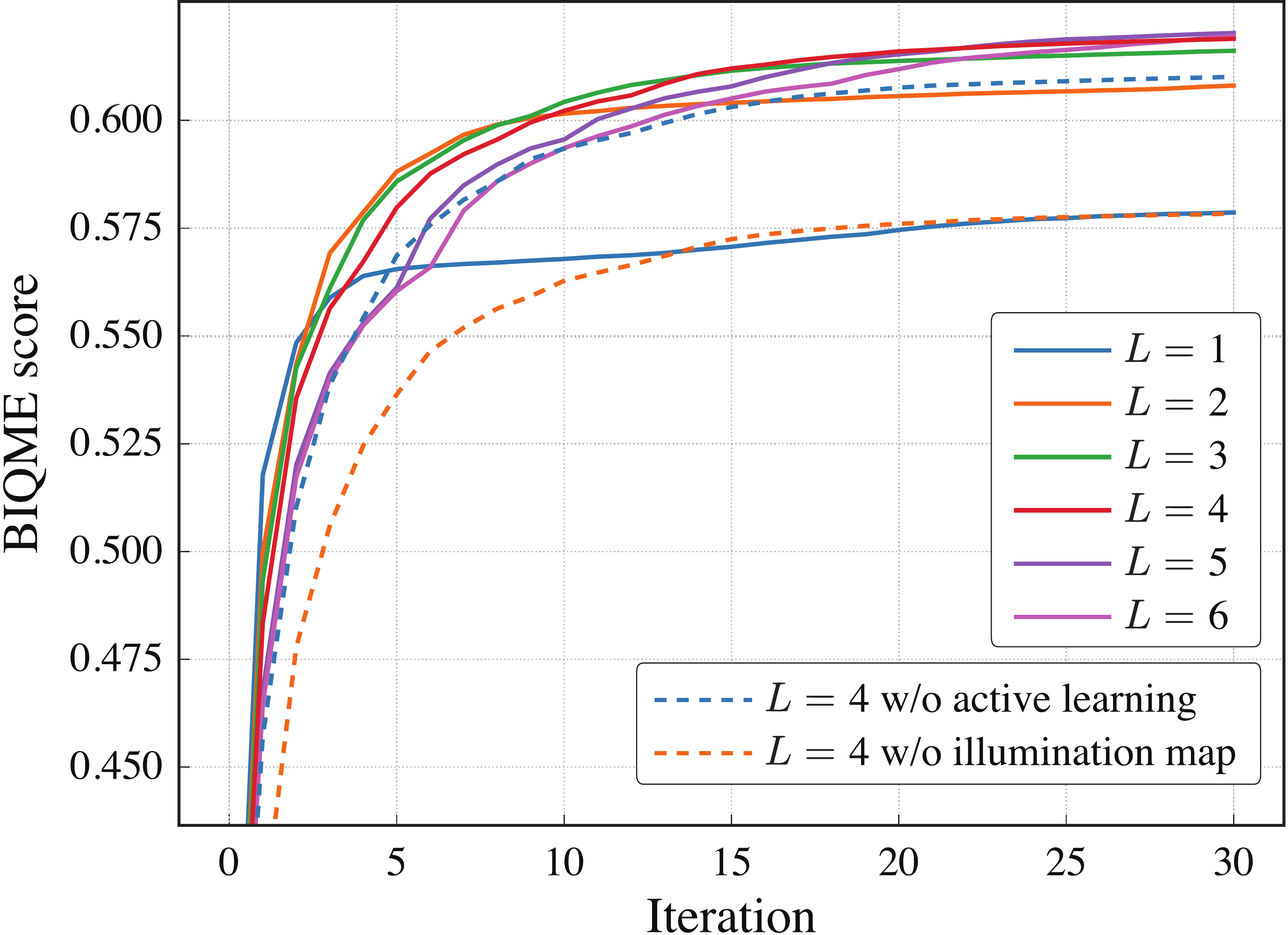}
\caption{Results when changing the number of pixels $L$ and
when not using active learning or the illumination map.}
\label{numpoints}
\end{figure}

\begin{figure}[t]
 \centering
 \includegraphics[width=1\hsize]{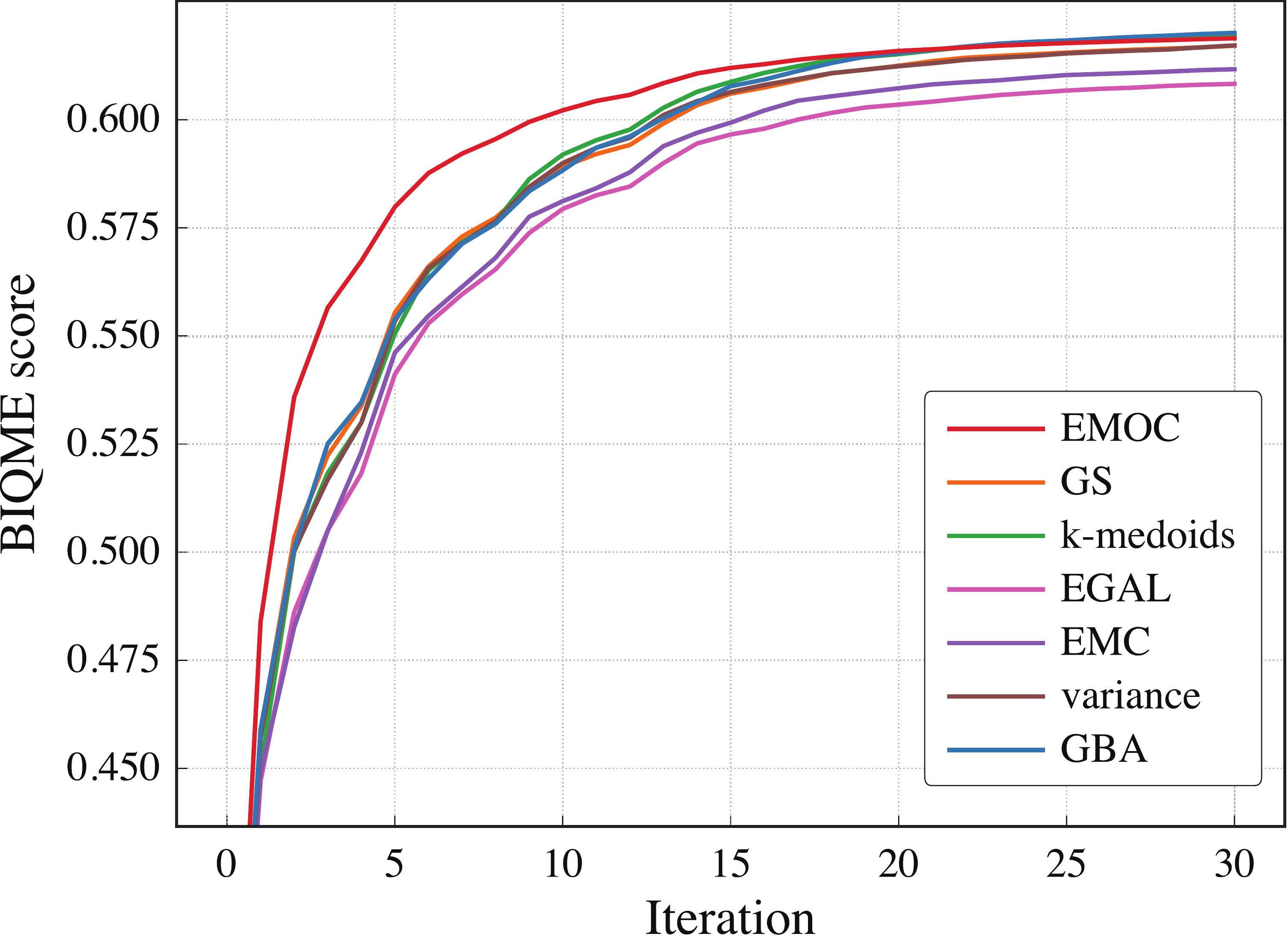}
\caption{Results using different active learning methods.}
\label{ALmethod}
\end{figure}

\begin{figure}[t]
 \includegraphics[width=1\hsize]{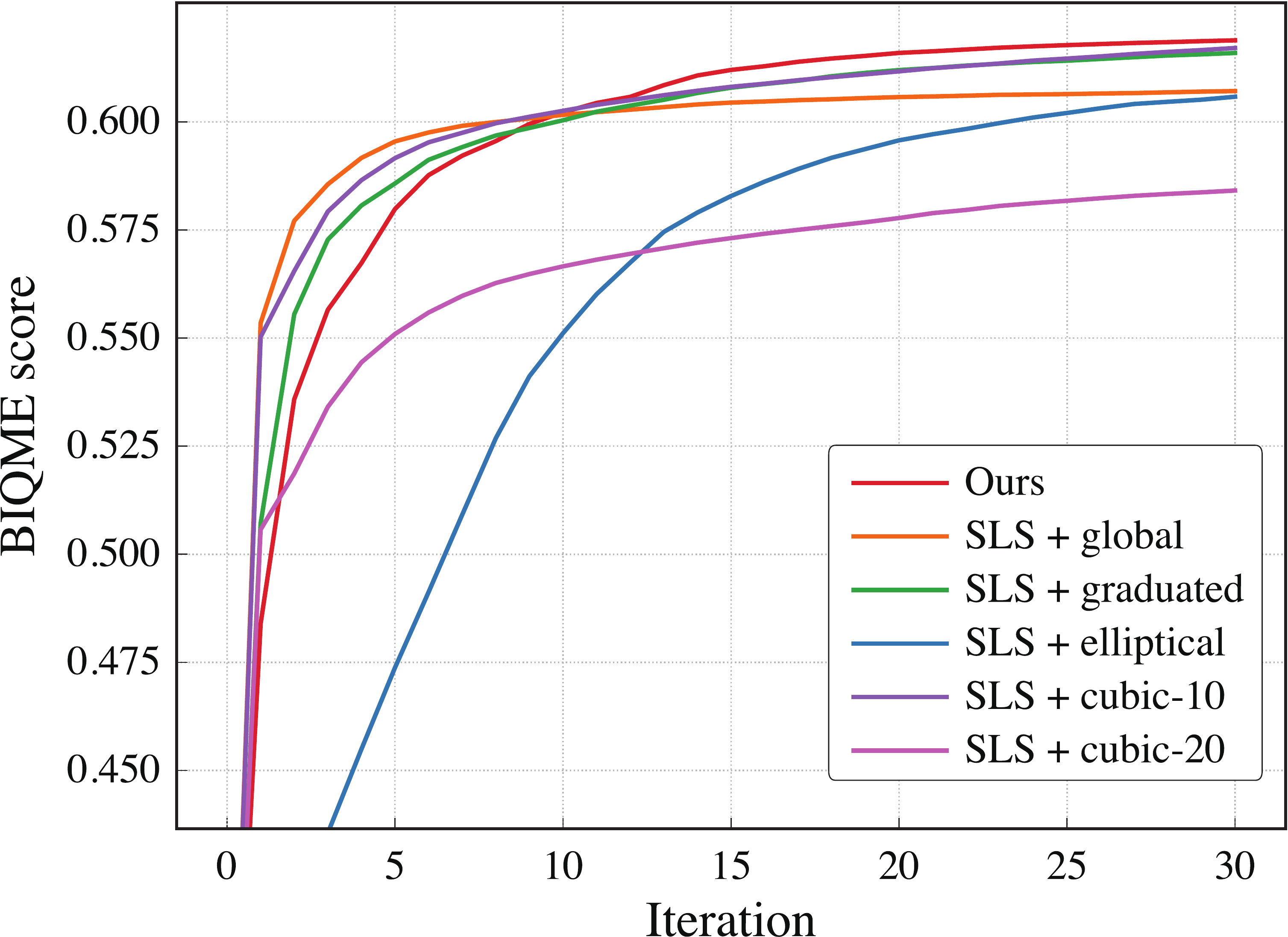}
\caption{Comparison with other filters~\cite{moran2020deeplpf}.}
\label{resultslocal}
\end{figure}

\begin{figure*}[t]
 \includegraphics[width=1\hsize]{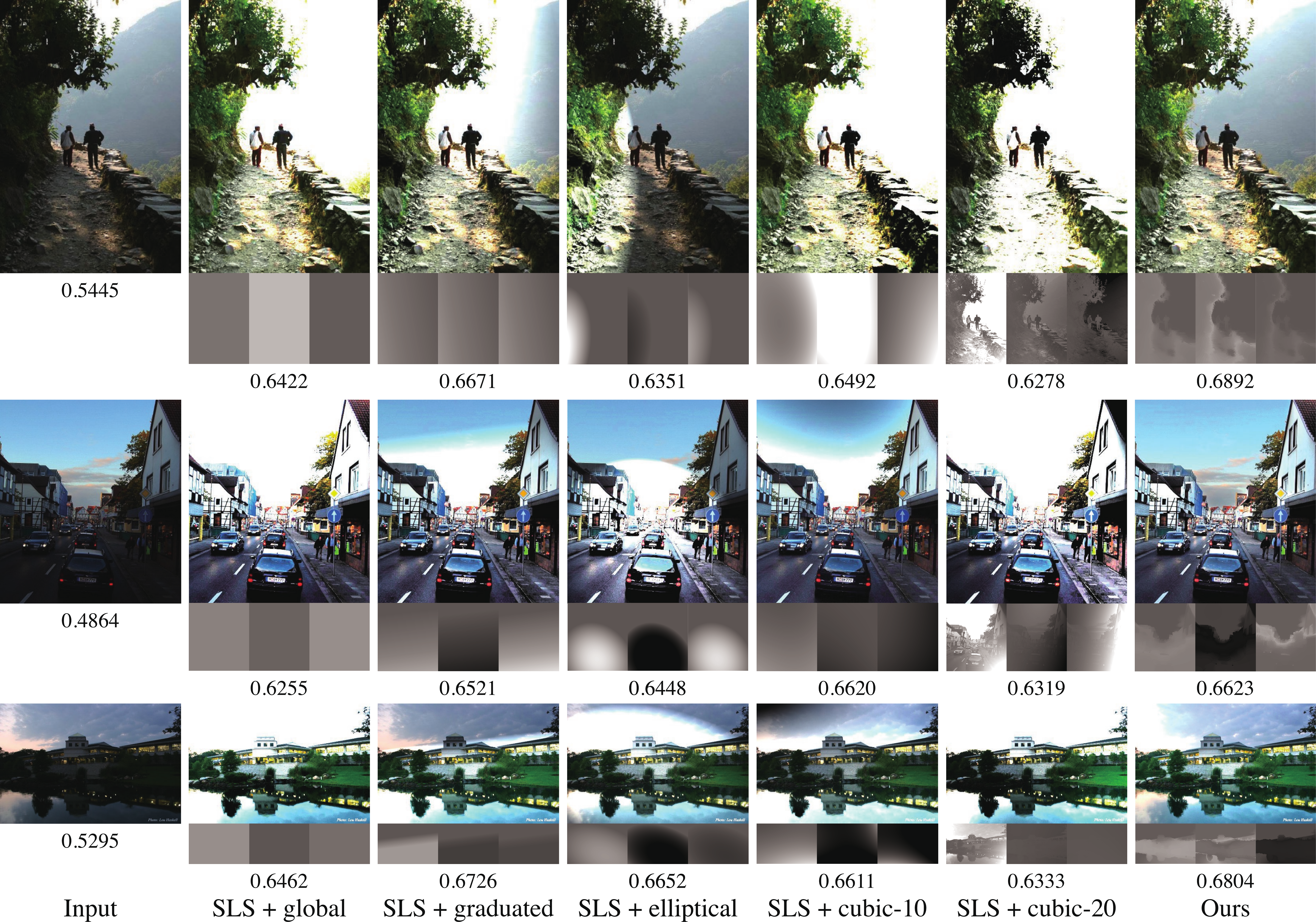}
\caption{Qualitative comparisons with other filters~\cite{moran2020deeplpf}. Each value denotes the BIQME scores.
Three images below each image are parameters for three image editing functions: Brightness, Saturation, and Contrast.
}
\label{resultsexplocal}
\end{figure*}

\section{Experiments}
\subsection{Evaluation of Our Local Filter}
To verify that the proposed filter is highly expressive and easy to optimize,
we conduct experiments using an image quality assessment model.
$\alpha^s_l$ in Eqs. (\ref{slider1}) and (\ref{slider2}) are adjusted to maximize
the quality score by the model,
and we evaluate the process of increasing the quality score.

\subsubsection{Dataset and Implementation}\label{hogehoge}

We use low-light images from the following datasets:
MEF~\cite{ma2015perceptual}, NPEA~\cite{wang2013naturalness}, LIME~\cite{guo2016lime}, DICM~\cite{lee2012contrast},
VV~\cite{vv_dataset}, and SICE~\cite{cai2018learning},
and the total number of images is 786.
SICE has multi-exposure sequences, and we use the darkest images in SICE.
As the image quality assessment model, we use BIQME~\cite{gu2017learning},
which evaluates images based on features such as brightness and colorfulness.
We use three image editing functions: Brightness, Saturation, and Contrast.
An exponential kernel is used as $\kappa$, and $r=1$.

\subsubsection{Number of Pixels}
We evaluate the relationship between the number of pixels $L$ and the performance.
As the number of pixels $L$ increases, the expressiveness becomes higher because the
GPR's prediction is improved,
but the optimization becomes more difficult because the iterations for the optimization, $L\times S$, increase.
Based on this property, we deduce the most efficient number of pixels.
The results of the optimization using different $L$ are shown in Figure~\ref{numpoints},
where we take the average BIQME score of the 786 images.
When $L=1$,
the convergence is fast, but the BIQME score does not increase well because of the lack of expressiveness.
The scores when $L=4$, $5$, and $6$ are almost the same at the 30th iteration,
and the convergence is fast when $L=4$;
therefore, we conclude that setting $L$ as 4 is the most efficient.
In the following experiments, we set $L$ as 4.

\subsubsection{Ablation Study}
Using active learning,
we select the key pixels which are the most suitable for predicting the other parameters.
Additionally, we achieve the edge-aware filter using the illumination map.
To verify whether these techniques actually contribute to the performance,
we conduct two ablation experiments: we select the key pixels randomly without active learning,
and use the filter without the illumination map.
As shown in the results in Figure~\ref{numpoints},
active learning and the illumination map
are necessary factors for the high performance.

\subsubsection{Active Learning Method}
To show that EMOC~\cite{kading2018active} is the best active learning method for our filter,
we replace EMOC with other label-independent active learning methods for regression:
GS~\cite{yu2010passive}, k-medoids~\cite{yu2010passive}, EGAL~\cite{hu2010egal},
EMC~\cite{cai2013maximizing}, variance~\cite{kading2018active}, and GBA~\cite{zhang2020graph},
and the results are shown in Figure~\ref{ALmethod}.
While the difference in scores between EMOC, GS, k-medoids, variance, and GBA is small at the 30th iteration,
EMOC outperforms other methods in early iterations;
therefore, we conclude that EMOC is the best for our purpose.

\begin{figure*}[t]
 \includegraphics[width=1\hsize]{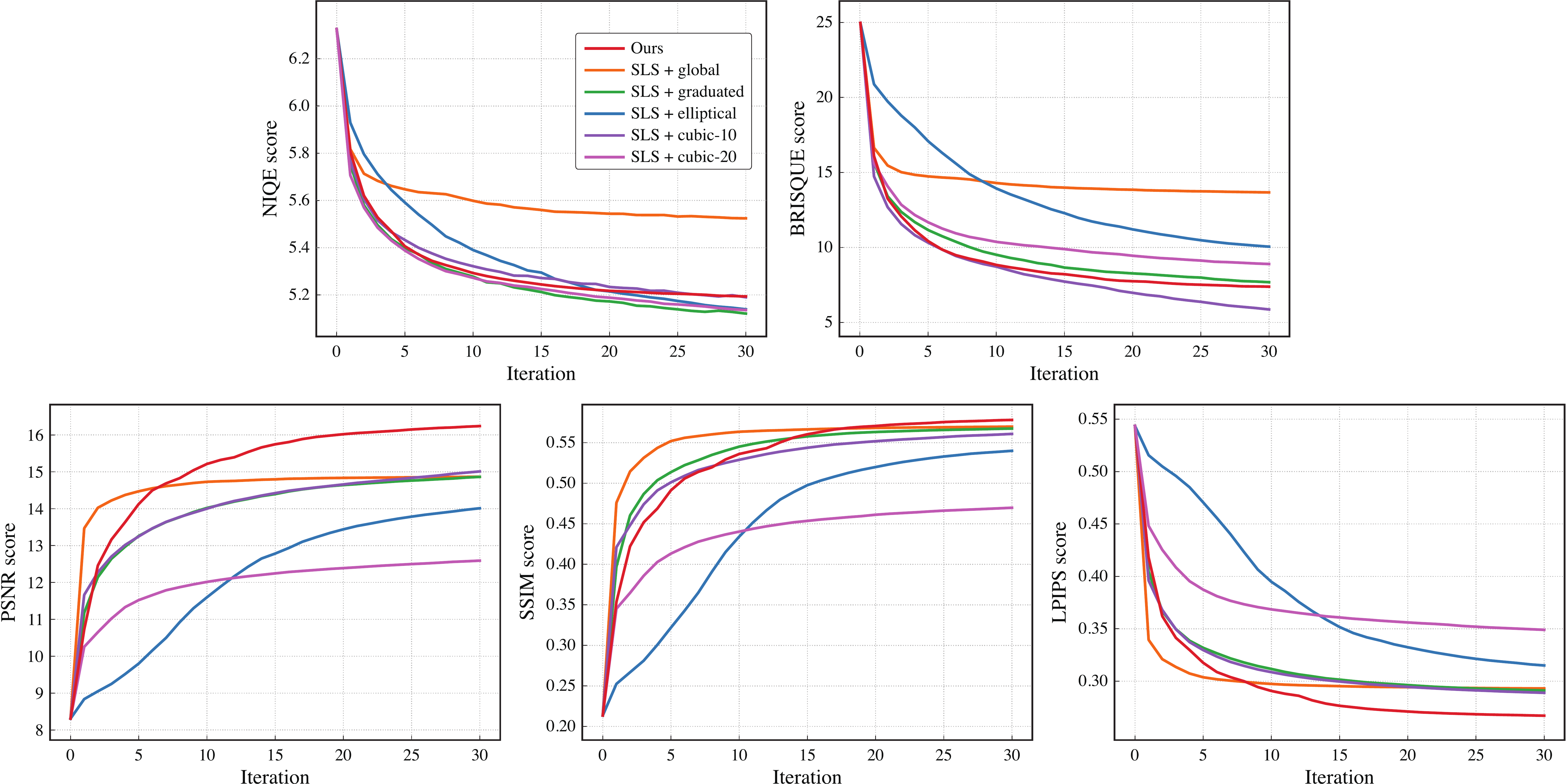}
\caption{
Comparisons with other filters~\cite{moran2020deeplpf} using five metrics:
NIQE~\cite{mittal2012making}, BRISQUE~\cite{mittal2012no}, PSNR, SSIM, and LPIPS~\cite{zhang2018unreasonable}.
}
\label{more_metrics}
\end{figure*}

\begin{figure}[t]
 \includegraphics[width=1\hsize]{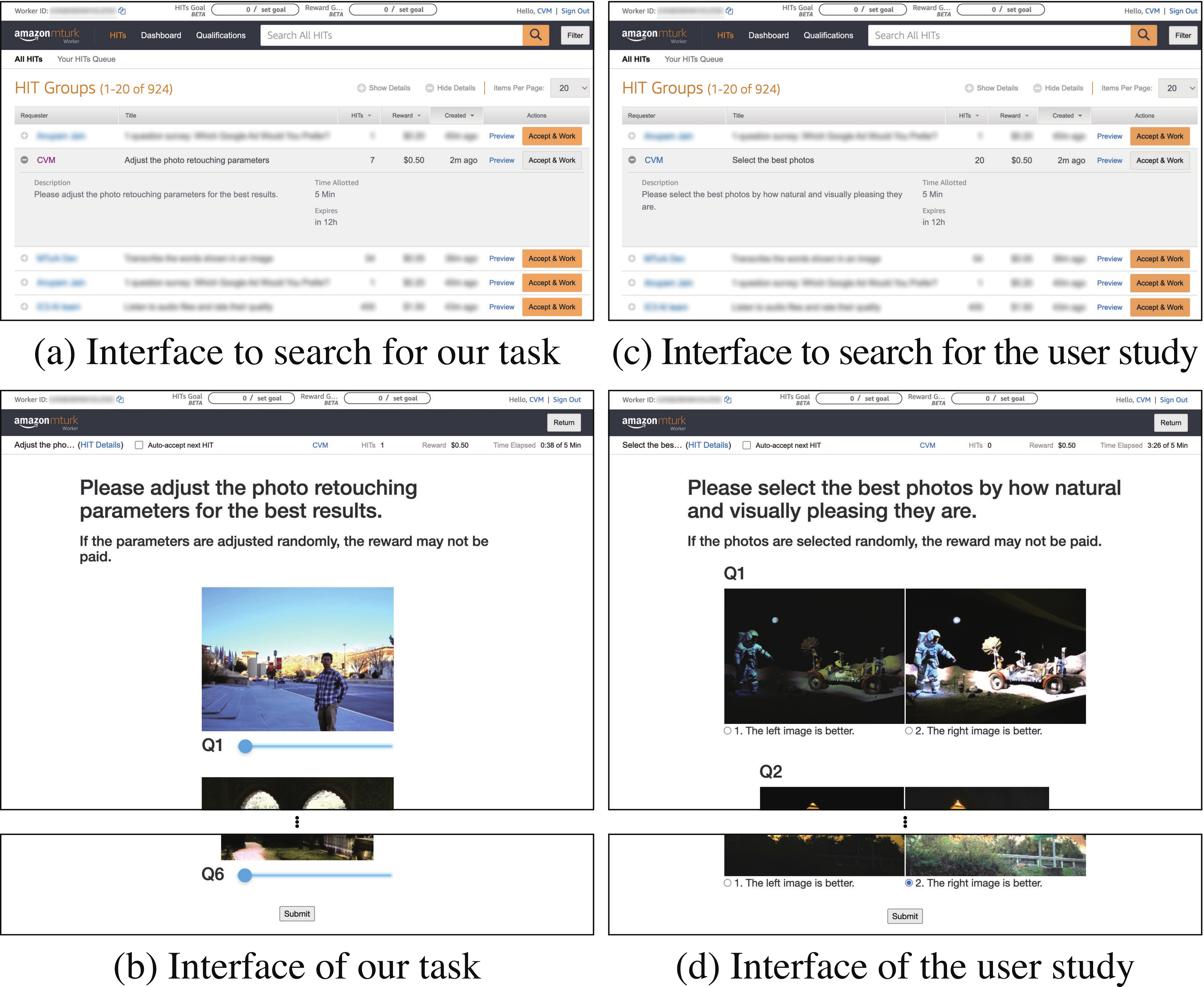}
\caption{Interface to the crowd workers in AMT. Best viewed in zoom.}
\label{interface}
\end{figure}

\begin{figure*}[p]
 \includegraphics[width=1\hsize]{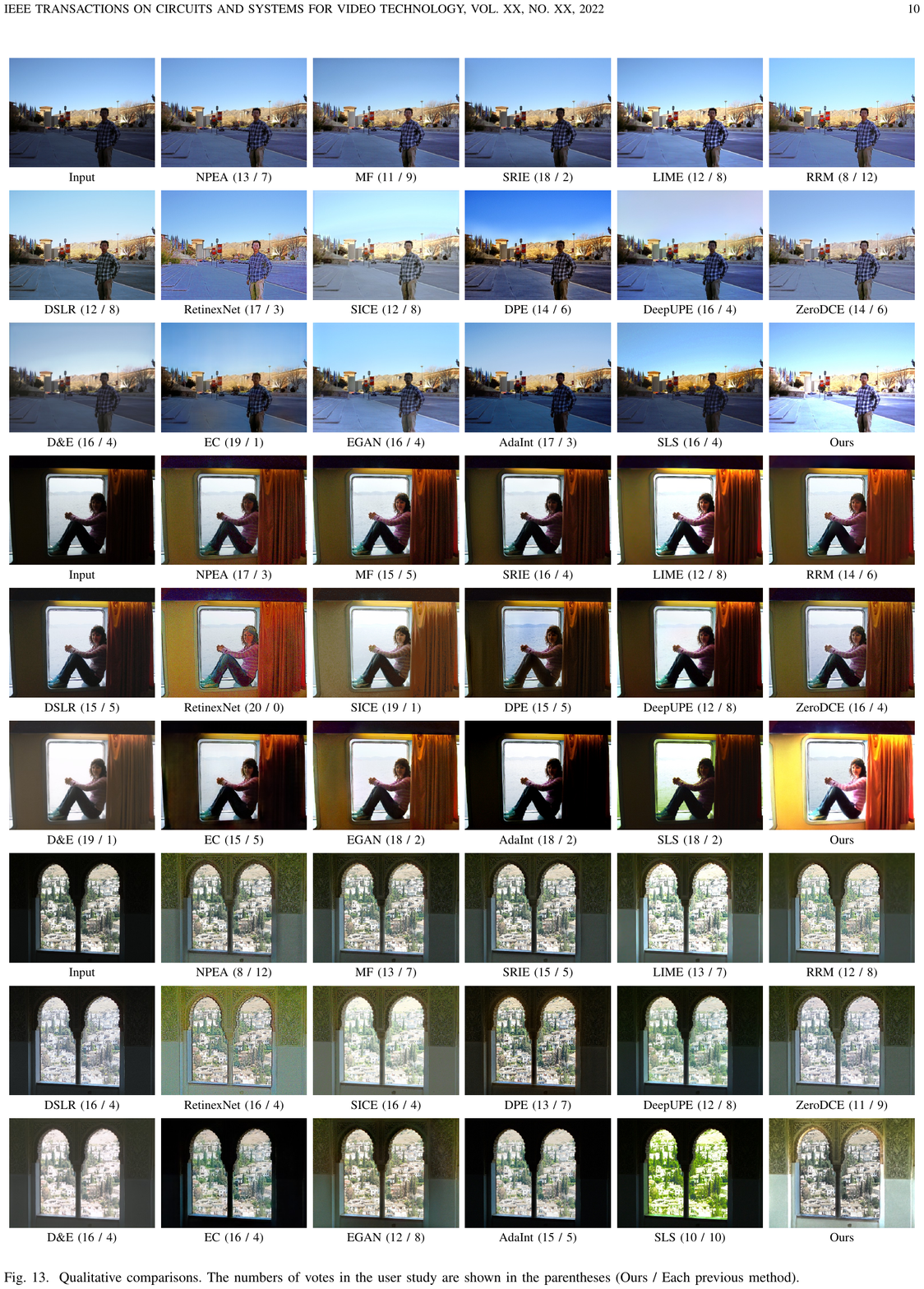}
 \caption{Qualitative comparisons.
 The numbers of votes in the user study are shown in the parentheses (Ours / Each previous method).}
\label{resultscrowd}
\end{figure*}

\subsubsection{Comparison with Existing Filters}
In DeepLPF~\cite{moran2020deeplpf}, four local parametric filters are proposed:
graduated filter, elliptical filter, and two polynomial filters (cubic-10 and cubic-20),
which are parameterized with eight, eight, 30, and 60 parameters, respectively.
The first three filters give
local effects depending on the xy-coordinates of the pixel,
whereas the last one gives a local edge-aware effect depending on the xy-coordinates
and the luminance value of the pixel.
To compare these filters with the proposed method,
we combine SLS~\cite{koyama2017sequential} with these local filters.
For example, when we use the graduated filter,
Eq. (\ref{tile_}) is replaced with ${\rm f}({\bm I}, {\bm P}) = {\rm f}({\bm I}, {\rm g^{graduated}}({\bm X}| \bm{p}^{\rm graduated}))$,
where $\bm{p}^{\rm graduated}\in\mathbb{R}^{8}$ determines the shape of the gradation,
and ${\rm g^{graduated}()}$ assigns parameters to ${\bm P}$ based on $\bm{p}^{\rm graduated}$.
$\bm{p}^{\rm graduated}$ is optimized using SLS.
In addition to the local filters, we use the global filter.
We show the results in Figure~\ref{resultslocal}.
These results show that the proposed filter can increase the BIQME score more efficiently
than the previous local filters.
Qualitative comparisons and optimized parameters are shown in Figure~\ref{resultsexplocal}.
Because the global filter cannot enhance images locally,
the global filter has to brighten the entire image to brighten the dark regions.
As a result, the result images are overexposed.
In the graduated filter, the centerline and width of the gradation are determined by the parameters.
In the elliptical filter, the center, semi-major axis, semi-minor axis, and rotation angle are determined by the parameters.
In the cubic-10, coefficients of a cubic polynomial are determined by the parameters,
where the xy-coordinates are used as variables.
Because these filters only depend on the xy-coordinates, the effects are not edge-aware.
In the cubic-20, the xy-coordinates and the luminance value are used as variables in the polynomial.
Cubic-20 is expected to give an edge-aware effect,
but the number of parameters is too large to effectively optimize the parameters.
As a result, the results by cubic-20 are overexposed.
Compared with these filters, the proposed filter is more expressive and easier to optimize.

For further comparisons, we evaluate each filter using various metrics.
We use two no-reference image quality assessment metrics (NIQE~\cite{mittal2012making} and BRISQUE~\cite{mittal2012no})
and three full-reference image quality assessment metrics (PSNR, SSIM, and LPIPS~\cite{zhang2018unreasonable}).
Higher PSNR and SSIM scores mean better results,
and lower NIQE, BRISQUE, and LPIPS scores mean better results.
As with BIQME, $\alpha^s_l$ in Eqs. (\ref{slider1}) and (\ref{slider2}) are adjusted to maximize or minimize
these metrics.
We use only the SICE dataset for full-reference metrics
because only the SICE dataset contains reference images.
We show the results in Figure~\ref{more_metrics}.
Among the six metrics including BIQME,
our filter achieves the best scores in four metrics: BIQME, PSNR, SSIM, and LPIPS,
and in the other two metrics, our filter achieves scores comparable to the best scores.
Therefore, we can conclude that our filter is more expressive and easier to optimize than previous filters.

\subsection{Evaluation of Our Enhancement}
To evaluate our content-aware local enhancement, we conduct experiments where
the slider is adjusted by real crowd workers.

\subsubsection{Implementation}
We use Amazon Mechanical Turk (AMT) as the crowdsourcing platform.
Because each crowd worker may respond with some “noise,”
our system deploys a microtask to seven crowd workers for each iteration.
Each microtask contains six pairs of an image and a slider;
five of them are for enhancement targets, and the last one is for a check task.
As the check task, we ask the crowd workers to adjust $\alpha_{\rm check}$ for the best result
and reject workers who set $\alpha_{\rm check}$ outside
the predefined range $[\alpha_{\rm check}^{\rm lower}, \alpha_{\rm check}^{\rm upper}]$,
because the result becomes too unnatural.
Our system deploys the same microtask again to the same number of workers as the rejected ones.
When seven responses are collected,
our system uses the median value of the seven responses as the best $\alpha^s_l$
and deploys a next microtask for the next iteration.
The next microtask is conducted by other seven workers generally.
To reduce the bias caused by the slider position,
the slider ends are reversed randomly for each worker.
We paid the workers 0.5 USD for each microtask
which includes the five enhancement targets and the one check task.
To save time and money in crowdsourcing, we apply an existing enhancement method,
LIME~\cite{guo2016lime}, to the input images before deploying tasks to the workers
as preprocessing.
We apply a denoising model~\cite{cheng2021nbnet} to the processed image
to reduce the noise caused by brightening the dark areas.
We set $S$ as 4.
Our enhancement process of the five enhancement targets is finished in about five hours.

For a more detailed explanation,
the interface to the crowd workers in AMT is shown in Figure~\ref{interface}.
In Figure~\ref{interface}(a), the workers search for tasks
based on the task title, description, and reward.
The title of our task is ``Adjust the photo retouching parameters'',
and the description is ``Please adjust the photo retouching parameters for the best results.''
Workers can see a preview of our task.
If a worker wants to participate in our task,
he/she presses the ``Accept \& Work'' button, which allows the worker to start our task.
Figure~\ref{interface}(b) shows the interface of our task.
At the top of our task, we write the instruction ``Please adjust the photo retouching parameters for the best results.''
In addition, we write ``If the parameters are adjusted randomly, the reward may not be paid.'' to make
the workers not to adjust the parameters randomly.
The workers adjust the sliders from Q1 to Q6;
when the all sliders are adjusted, the workers press ``Submit'' button to finish our task.
The reward for the workers is automatically paid.
We do not set any conditions for workers to participate in our task.

\subsubsection{Qualitative Evaluation}
We conduct a qualitative comparison with the existing methods.
For comparison, we use five rule-based methods
(NPEA~\cite{wang2013naturalness}, MF~\cite{fu2016fusion}, SRIE~\cite{fu2016weighted},
LIME~\cite{guo2016lime}, and RRM~\cite{li2018structure});
10 learning methods (DSLR~\cite{ignatov2017dslr}, RetinexNet~\cite{wei2018deep}, SICE~\cite{cai2018learning},
DPE~\cite{chen2018deep}, DeepUPE~\cite{wang2019underexposed}, ZeroDCE~\cite{guo2020zero},
D\&E~\cite{xu2020learning}, EC~\cite{afifi2021learning},
EGAN~\cite{jiang2021enlightengan}, and AdaInt~\cite{yang2022adaint});
and one crowd-powered method (SLS~\cite{koyama2017sequential}).
Because there are many learning methods~\cite{yan2016automatic,gharbi2017deep,wang2019underexposed,moran2020deeplpf,kim2020global,he2020conditional,afifi2021learning,kim2021representative,zhao2021deep,song2021starenhancer,wang2021real,yang2022adaint,zhang2021star}
using the MIT-Adobe 5K dataset~\cite{bychkovsky2011learning},
we use a representative method (DeepUPE~\cite{wang2019underexposed}), a method which
proposes augmentation for the dataset (EC~\cite{afifi2021learning}),
and one of the best-performing methods (AdaInt~\cite{yang2022adaint}).
We use the official implementation except for SLS.
For SLS, because only the function $\rm SLS()$ is publicly available,
we create the interface for AMT.

We show the qualitative comparisons in Figure~\ref{resultscrowd}.
The rule-based methods (NPEA, MF, SRIE, LIME, and RRM) locally brighten the dark areas,
but the dark areas remain under-exposed.
These methods cannot apply different effects to each image considering the contents,
which results in low quality.
DSLR and D\&E do not enhance the images locally
because the reference images in the training datasets
({\it i.e.}, the DPED~\cite{ignatov2017dslr} and the sRGB-SID dataset~\cite{xu2020learning}, respectively)
are created by changing the shooting condition globally.
Similarly, DeepUPE, EC, and AdaInt do not enhance the images locally
because the reference images in the training dataset
({\it i.e.}, the MIT-Adobe 5K dataset~\cite{bychkovsky2011learning}) are retouched globally.
RetinexNet achieves local enhancement using a model inspired by rule-based methods
but generates slightly unnatural images, which is a limitation of the model.
SICE also generates locally enhanced but slightly unnatural results,
because the used reference images contain blur and ghosting artifacts.
DPE, ZeroDCE, and EGAN train the models without paired datasets and enhance the images locally,
but the quality is low because they are inferior to paired learning methods.
The crowd-powered method, SLS, does not enhance the images locally because only a global filter is used.
Compared to these methods,
the proposed method brightens the dark regions appropriately for all the images,
and for the bottom image,
the outside of the window is overexposed in LIME, SICE, and D\&E
but is properly exposed in the proposed method.
These show the advantage of our content-aware local enhancement.

\begin{figure}[t]
 \includegraphics[width=1\hsize]{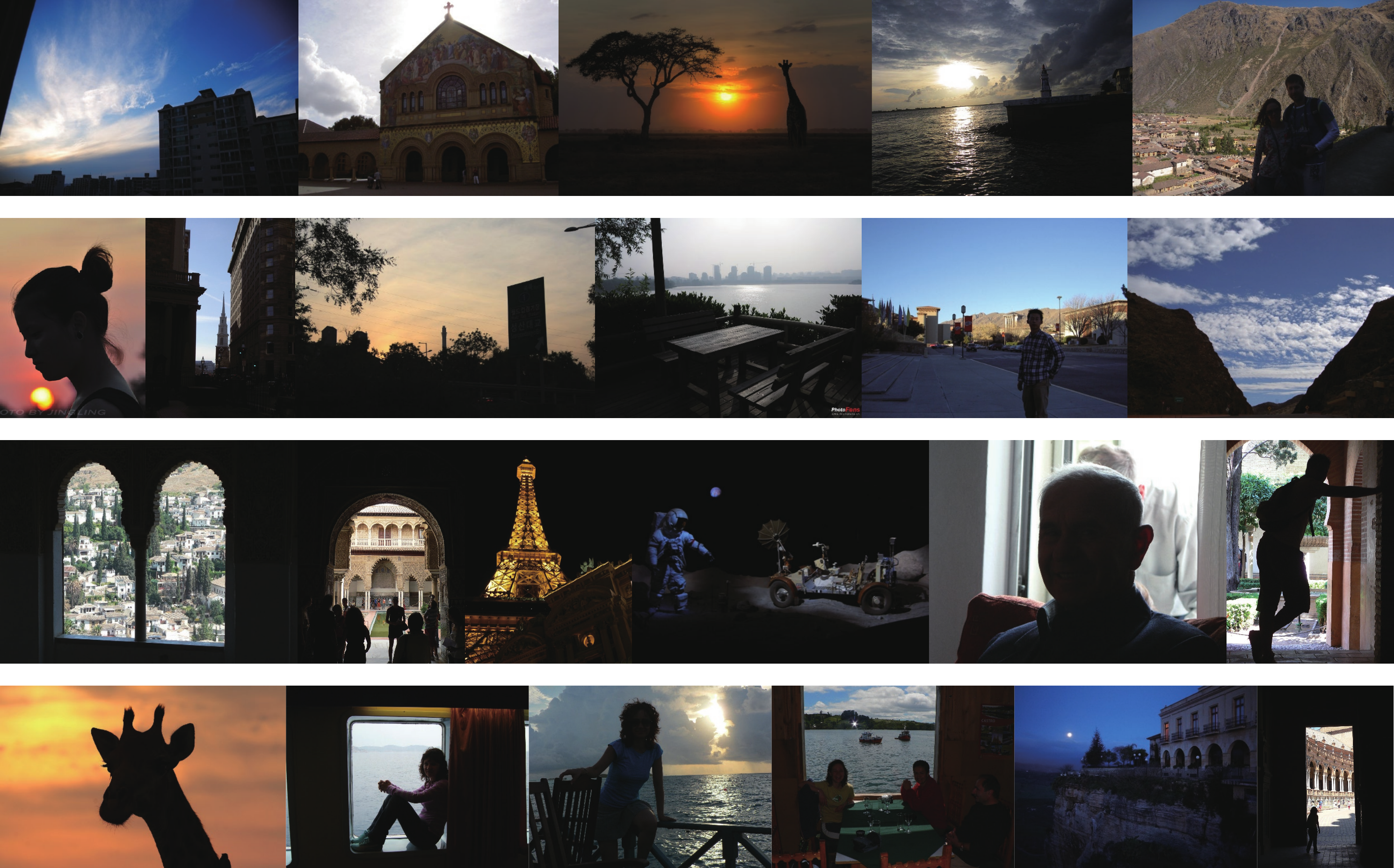}
\caption{23 images used in the user study.}
\label{thumbnail}
\end{figure}

\begin{table}[!t]
  \caption{Result of the user study.}
\centering
\begin{tabular}{l|c}
\hline
\rule[0mm]{0mm}{3.5mm}Method & Average voting rate [\%]\\
\hline\hline
  \rule[0mm]{0mm}{3mm}Ours / NPEA~\cite{wang2013naturalness} & {\bf 69.8} / 30.2\\
  \rule[0mm]{0mm}{3mm}Ours / MF~\cite{fu2016fusion} & {\bf 63.3} / 36.7\\
  \rule[0mm]{0mm}{3mm}Ours / SRIE~\cite{fu2016weighted} & {\bf 74.3} / 25.7\\
  \rule[0mm]{0mm}{3mm}Ours / LIME~\cite{guo2016lime} & {\bf 56.7} / 43.3\\
  \rule[0mm]{0mm}{3mm}Ours / RRM~\cite{li2018structure} & {\bf 61.1} / 38.9\\
  \rule[0mm]{0mm}{3mm}Ours / DSLR~\cite{ignatov2017dslr} & {\bf 75.0} / 25.0\\
  \rule[0mm]{0mm}{3mm}Ours / RetinexNet~\cite{wei2018deep} & {\bf 78.3} / 21.7\\
  \rule[0mm]{0mm}{3mm}Ours / SICE~\cite{cai2018learning} & {\bf 67.2} / 32.8\\
  \rule[0mm]{0mm}{3mm}Ours / DPE~\cite{chen2018deep} & {\bf 73.7} / 26.3\\
  \rule[0mm]{0mm}{3mm}Ours / DeepUPE~\cite{wang2019underexposed} & {\bf 69.6} / 30.4\\
  \rule[0mm]{0mm}{3mm}Ours / ZeroDCE~\cite{guo2020zero} & {\bf 67.0} / 33.0\\
  \rule[0mm]{0mm}{3mm}Ours / D\&E~\cite{xu2020learning} & {\bf 79.6} / 20.4\\
  \rule[0mm]{0mm}{3mm}Ours / EC~\cite{afifi2021learning} & {\bf 82.8} / 17.2\\
  \rule[0mm]{0mm}{3mm}Ours / EGAN~\cite{jiang2021enlightengan} & {\bf 70.7} / 29.3\\
  \rule[0mm]{0mm}{3mm}Ours / AdaInt~\cite{yang2022adaint} & {\bf 78.7} / 21.3\\
  \rule[0mm]{0mm}{3mm}Ours / SLS~\cite{koyama2017sequential} & {\bf 84.1} / 15.9\\
\hline
\end{tabular}
\label{resultsuserstudy}
\end{table}

\begin{table}[!t]
  \caption{Quantitative comparisons with other enhancement methods.}
\centering
{\tabcolsep=0.88mm
\begin{tabular}{l|ccc|ccc}
\hline
\rule[0mm]{0mm}{3.5mm}Method & BIQME$\uparrow$ & NIQE$\downarrow$ & BRISQUE$\downarrow$ & PSNR$\uparrow$ & SSIM$\uparrow$ & LPIPS$\downarrow$\\
\hline\hline
\rule[0mm]{0mm}{3mm}NPEA~\cite{wang2013naturalness}	&	0.5745	&	4.757	&	19.26	 & 12.44 & 0.4107 & {\bf 0.4091}\\
\rule[0mm]{0mm}{3mm}MF~\cite{fu2016fusion}	&	0.5611	&	4.542	&	17.37	 & 11.75 & 0.3941 & 0.4150\\
\rule[0mm]{0mm}{3mm}SRIE~\cite{fu2016weighted}	&	0.5440	&	4.646	&	19.97	 & 9.46 & 0.3054 & 0.4678\\
\rule[0mm]{0mm}{3mm}LIME~\cite{guo2016lime} 	&	0.5979	&	4.828	&	27.76	 & 13.23 & 0.4408 & 0.4400\\
\rule[0mm]{0mm}{3mm}RRM~\cite{li2018structure}	&	0.5978	&	5.322	&	30.26	 & 10.91 & 0.3575 & 0.5398\\
\rule[0mm]{0mm}{3mm}DSLR~\cite{ignatov2017dslr}	&	0.5177	&	5.274	&	27.72	 & 11.80 & 0.4010 & 0.4872\\
\rule[0mm]{0mm}{3mm}RetinexNet~\cite{wei2018deep}	&	0.5657	&	5.428	&	30.68	 & 13.18 & 0.4361 & 0.4551\\
\rule[0mm]{0mm}{3mm}SICE~\cite{cai2018learning}	&	0.5559	&	4.320	&	27.11	 & 13.23 & 0.4500 & 0.5016\\
\rule[0mm]{0mm}{3mm}DPE~\cite{chen2018deep}	&	0.5315	&	4.085	&	13.85	 & 8.79 & 0.2897 & 0.5371\\
\rule[0mm]{0mm}{3mm}DeepUPE~\cite{wang2019underexposed}	&	0.5452	&	4.642	&	13.33	 & 10.00 & 0.2967 & 0.5114\\
\rule[0mm]{0mm}{3mm}ZeroDCE~\cite{guo2020zero}	&	0.5678	&	4.634	&	20.12	 & 11.39 & 0.3872 & 0.4107\\
\rule[0mm]{0mm}{3mm}D\&E~\cite{xu2020learning}	&	0.5955	&	4.826	&	30.71	 & 13.24 & 0.4277 & 0.5430\\
\rule[0mm]{0mm}{3mm}EC~\cite{afifi2021learning}	&	0.5315	&	4.085	&	13.85	 & 11.25 & 0.3387 & 0.6067\\
\rule[0mm]{0mm}{3mm}EGAN~\cite{jiang2021enlightengan}	&	0.5919	&	3.933	&	14.10	 & 12.27 & 0.4062 & 0.4502\\
\rule[0mm]{0mm}{3mm}AdaInt~\cite{yang2022adaint}	&	0.5066	&	5.444	&	21.04	 & 9.23 & 0.2455 & 0.5723\\
\rule[0mm]{0mm}{3mm}SLS~\cite{koyama2017sequential}	&	0.4858	&	4.998	&	{\bf 12.99}	& 7.98 & 0.2004 & 0.6843\\
\rule[0mm]{0mm}{3mm}Ours	&	{\bf 0.6162}	&	{\bf 3.764}	&	15.95	 & {\bf 13.72} & {\bf 0.4589} & 0.4909\\
\hline
\end{tabular}
}
\label{quantitative}
\end{table}

\subsubsection{User Study}
We evaluate the proposed method through a user study.
We randomly select 23 images (which is the same number as that used in EGAN)
and perform enhancement using each existing method and the proposed method.
The 23 images are shown in Figure~\ref{thumbnail}.
20 crowd workers via AMT are asked to compare two images
enhanced by our method and one of the previous methods;
they are then instructed to select the better image.
All images are arranged randomly to avoid bias.
Table~\ref{resultsuserstudy} lists the average voting rate.
Our proposed method achieves a higher rate than all the
existing methods, which shows that it is capable of high-quality enhancement.

We show the interface to search for the user study in Figure~\ref{interface}(c).
We set the title and description
as ``Select the best photos'' and ``Please select the best photos by how natural and visually pleasing they are,''
respectively.
We show the interface of our user study in Figure~\ref{interface}(d).
At the top of our user study, we write the instruction ``Please select the best photos by how natural and visually pleasing they are.''
Crowd workers select the best photos;
when all the questions are answered, the workers press ``Submit'' button to finish our user study.
The reward for the workers is automatically paid.
We do not set any conditions for workers to participate in our user study.

\subsubsection{Quantitative Evaluation}
We evaluate the proposed method using three no-reference image quality assessment metrics
(BIQME~\cite{gu2017learning}, NIQE~\cite{mittal2012making}, and BRISQUE~\cite{mittal2012no})
and three full-reference image quality assessment metrics (PSNR, SSIM, and LPIPS~\cite{zhang2018unreasonable}).
Higher BIQME, PSNR, and SSIM scores mean better results,
and lower NIQE, BRISQUE, and LPIPS scores mean better results.
We use the 23 images used in the user study for the no-reference image quality assessment metrics
and randomly select 20 images from the SICE dataset~\cite{cai2018learning} for the full-reference image quality assessment metrics.
We show the average scores in Table~\ref{quantitative}.
Among the six metrics, our method achieves the best scores in four metrics: BIQME, NIQE, PSNR,
and SSIM, and in the other two metrics, our method
achieves scores comparable to the best scores.
Therefore, we can conclude that our method is quantitatively superior to the previous methods.

\begin{figure}[t]
	\includegraphics[width=1\hsize]{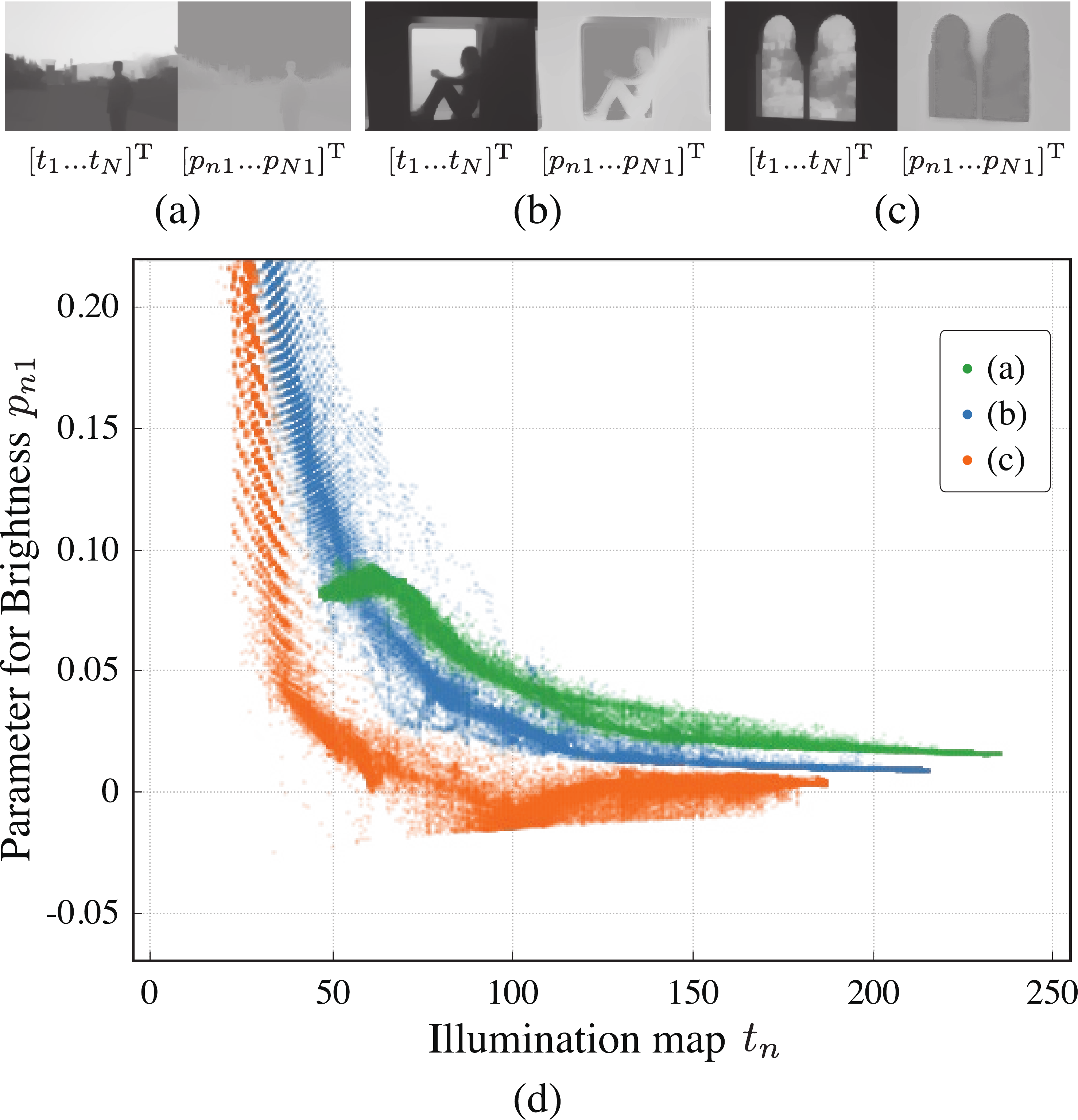}
\caption{Relationship between the illumination map ${\bm t} = [t_1...t_N]^{\rm T}$ and the parameters for Brightness $[p_{n1}...p_{N1}]^{\rm T}$.}
\label{contentaware}
\end{figure}

\subsubsection{Visualization of Content-awareness}
To verify that the proposed method is content-aware rather than just illumination-aware,
we show the illumination map ${\bm t} = [t_1...t_N]^{\rm T}$ and the parameters for Brightness $[p_{n1}...p_{N1}]^{\rm T}$
(Figures~\ref{contentaware}(a), (b), and (c))
and plot $\{(t_n, p_{n1})\}_{n=1}^N$ (Figure~\ref{contentaware}(d)).
Note that ${p_{n1}}$ is not $\bm{p}_{n_1}$ but the first element of $\bm{p}_{n}$.
When $p_{n1}=0$, no effects are applied to the images;
when $p_{n1}$ is positive, the images become brighter,
and when $p_{n1}$ is negative, the images become darker.
If the crowd workers determine the parameters based only on the illumination map,
the distributions of $(t_n, p_{n1})$ should be consistent for all the images;
however, the distributions are different.
For example, when $t_n>100$,
$p_{n1}$ in Figures~\ref{contentaware}(a) and (c) takes positive values and
almost zero values, respectively.
This is because
the pixels where $t_n>100$ in Figures~\ref{contentaware}(a) and (c) are the sky area and the building area, respectively,
and the crowd workers consider that the sky should be brighter, and
the buildings should remain the details to be more visually pleasing.
These prove that the proposed method is content-aware rather than just illumination-aware.

\subsubsection{Robustness of Our Enhancement}
\begin{figure}[t]
	\includegraphics[width=1\hsize]{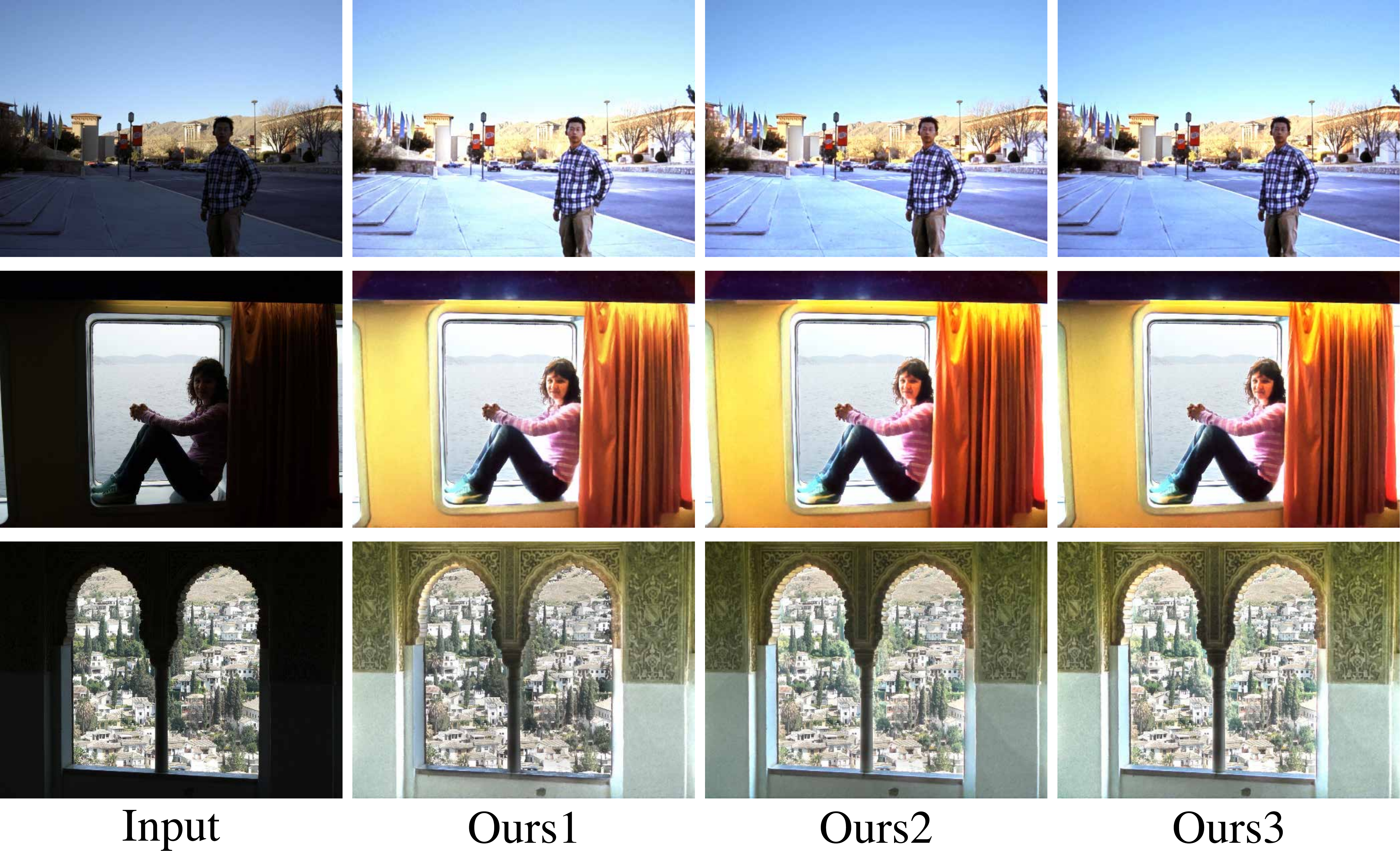}
\caption{The results of our enhancement method from three trials.}
\label{robustness}
\end{figure}

Our enhancement method outputs different results for each trial.
This is because the initial parameters $\bm{p}^1_{n_l}$ and $\bm{\bar p}^{1}_{n_l}$
are randomly determined,
and the crowd workers do not always return the same responses.
To verify the robustness of our enhancement,
we conduct the experiment three times,
and the results are shown in Figure~\ref{robustness}.
We can observe slight differences among the results,
but they are within a negligible range.
Besides, they are all visually pleasing.

\begin{figure}[t]
	\includegraphics[width=1\hsize]{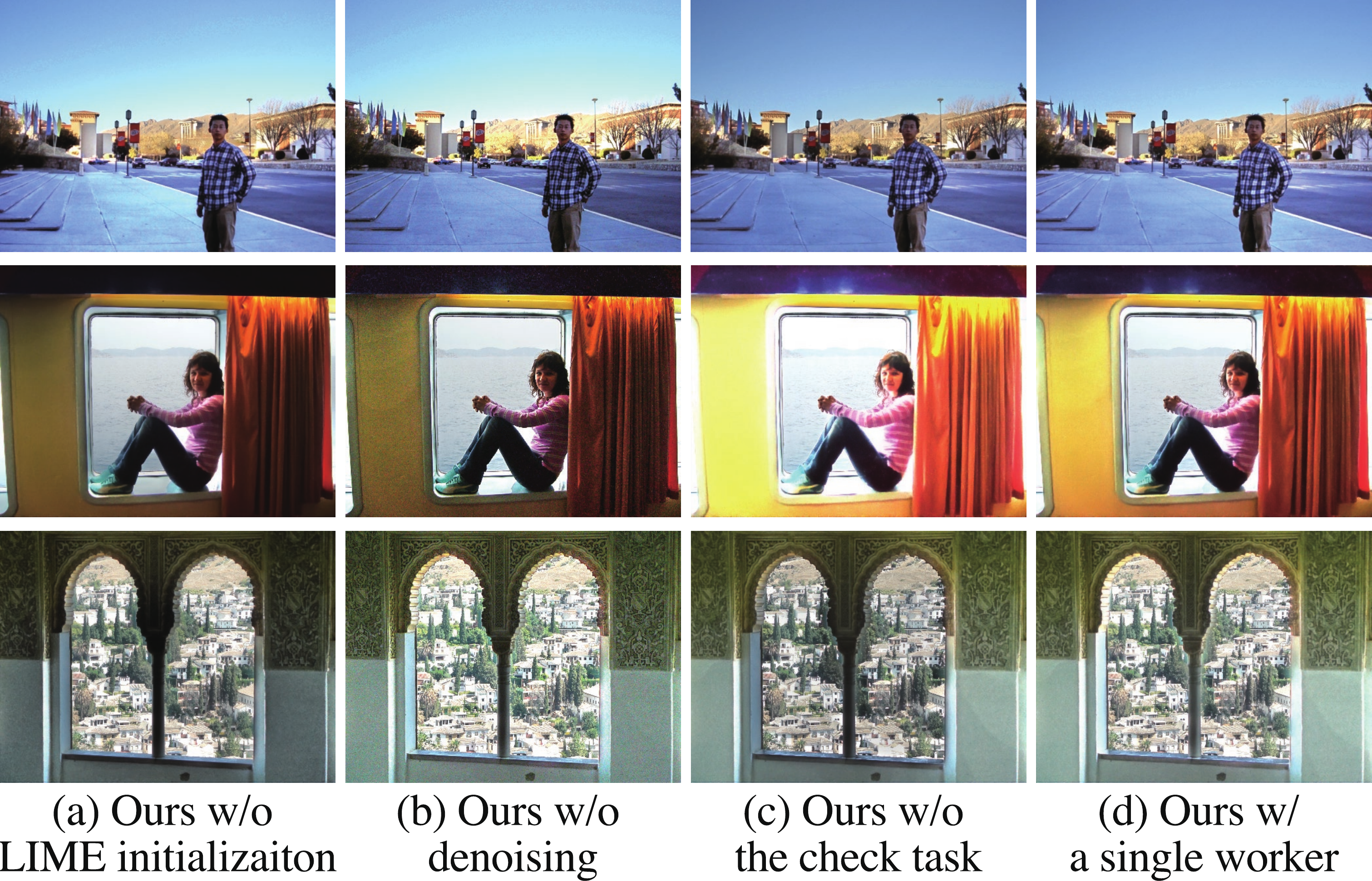}
\caption{The results of the ablation studies.}
\label{crowd_ablation}
\end{figure}

\subsubsection{Ablation Study}
To save time and money in crowdsourcing, we apply an existing enhancement method,
LIME~\cite{guo2016lime}, to the input images before deploying the tasks to the workers
as initialization, and
we apply a denoising model~\cite{cheng2021nbnet} to the processed image
to reduce the noise caused by brightening the dark areas.
To make our method robust,
we set the check task and deploy the tasks to seven crowd workers.
To verify whether these techniques actually
contribute to the performance, we conduct ablation experiments
and show the results in Figure~\ref{crowd_ablation}.
When we do not apply LIME to the input images (Figure~\ref{crowd_ablation}(a)),
the middle image is underexposed because the parameter optimization needs more iterations.
When we do not use the denoising model (Figure~\ref{crowd_ablation}(b)),
the middle image is underexposed because the workers avoid noise amplification.
When we do not set the check task (Figure~\ref{crowd_ablation}(c)),
the top image is underexposed, and the middle image is overexposed;
when we deploy the tasks to a single worker (Figure~\ref{crowd_ablation}(d)),
the top image is underexposed,
which means that these two techniques contribute to the robustness.

\section{Conclusions}
In the existing studies for photo enhancement,
rule-based methods, learning methods, and crowd-powered methods are available,
but the results enhanced by these methods are of low quality
because they are either not content-aware or not local.
To achieve content-aware local enhancement,
we proposed the crowd-powered local enhancement method,
where crowd workers locally optimize the parameters for image editing functions.
To make it easier to locally optimize the parameters, we proposed a novel active learning based local filter.
Only the parameters at a few key pixels need to be determined,
and the parameters at the other pixels are automatically predicted using the regression model.
We used active learning to select the key pixels that are the most suitable for predicting the other parameters,
and we used the illumination map to achieve edge-aware enhancement.
We broke down the task of optimizing multiple parameters into a sequence of single slider manipulation,
and crowd workers only need to adjust the single slider multiple times so that the image looks the best.
In the experiment using the image quality assessment model, BIQME,
our filter improved BIQME scores more efficiently than our filter without active learning or without the illumination map,
and our filter outperformed the previous local filters.
In the experiment using the real crowdsourcing,
our method generated more visually pleasing results than the results of the previous rule-based methods,
learning methods, and crowd-powered methods,
and our method was most highly evaluated in the user study.
Based on these results, we can conclude that
our active learning based local filter is highly expressive and easy to optimize,
and our content-aware local enhancement method can achieve higher-quality enhancement.

{\bf Limitation} The disadvantage of the proposed method is
the long processing time.
While existing rule-based and learning methods can enhance a single image in a few seconds at the latest,
the proposed method takes several hours to enhance a single image.
After our task is deployed, we need to wait a few dozen minutes until seven workers find and join our task;
therefore, such long processing time is required.
The proposed method is useful only to those who want to make enhanced results as visually pleasing as possible
without worrying about processing time.

{\bf Future work} In future work,
it is possible to create a novel dataset using the proposed method.
In learning methods, paired datasets of low-quality original images and high-quality reference images
are necessary to train models,
but locally-edited high-quality paired datasets do not exist because they are expensive to create.
We can easily obtain locally-edited results by using the proposed method,
which makes it easier to create a locally-edited paired dataset.
We will be able to achieve a content-aware local enhancement model
by training an enhancement model using the novel dataset,
and the problem of the long processing time will be solved.

\newpage

   \begin{IEEEbiography}[{\includegraphics[width=1in,height=1.25in,clip,keepaspectratio]{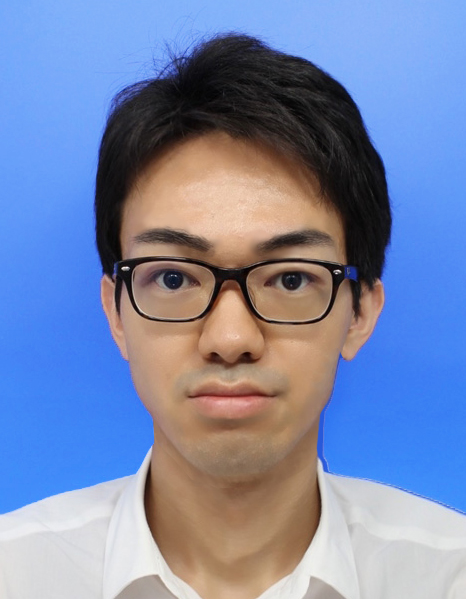}}]{Satoshi Kosugi}
     received the B.E.
and M.S. degrees in information and communication
engineering from The University of Tokyo, Japan,
in 2018 and 2020, respectively, where he is currently
pursuing the Ph.D. degree. His research interests
include computer vision, with particular interest in
image enhancement.
   \end{IEEEbiography}

   \begin{IEEEbiography}[{\includegraphics[width=1in,height=1.25in,clip,keepaspectratio]{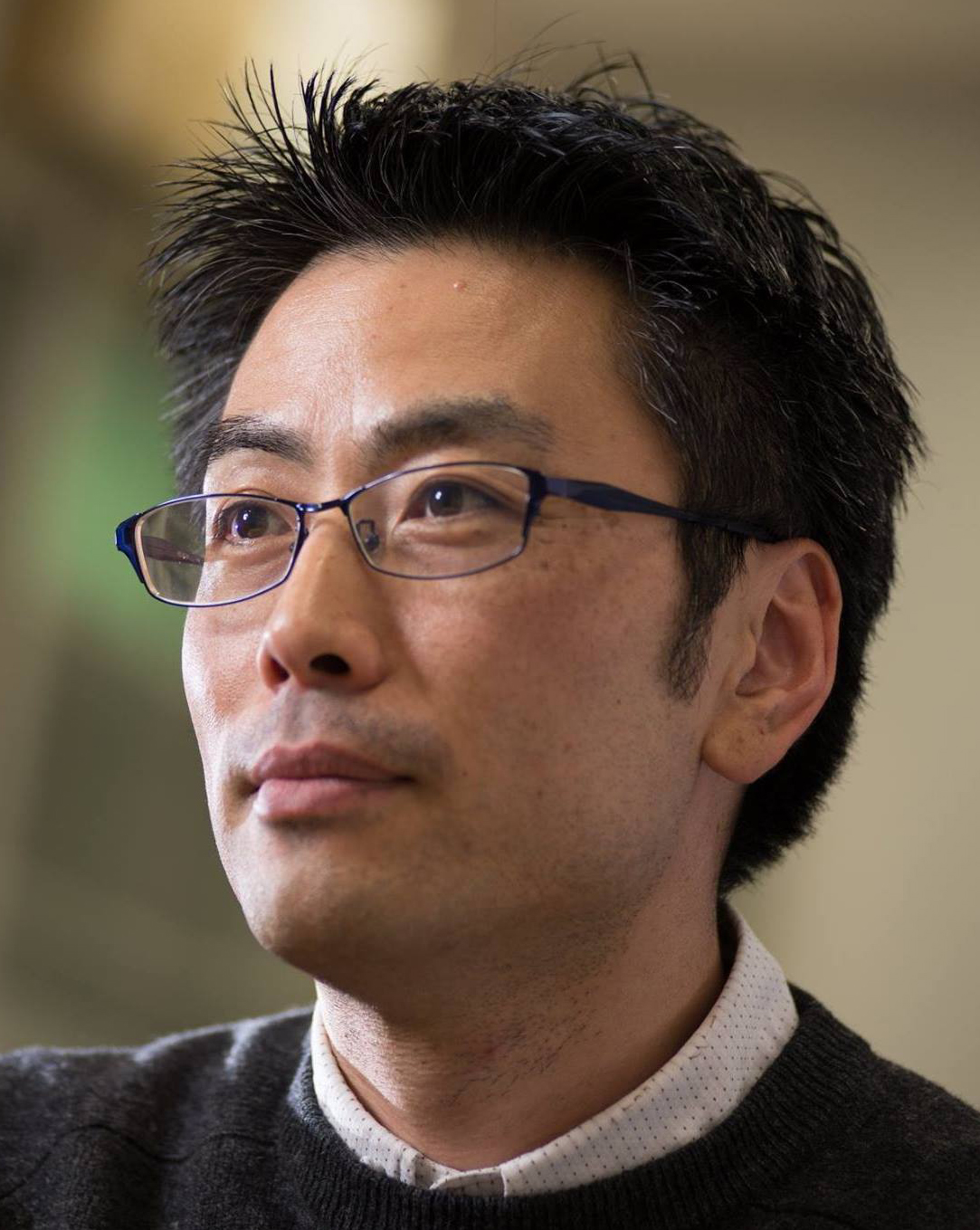}}]{Toshihiko Yamasaki}
   (Member, IEEE) received the
B.S., M.S., and Ph.D. degrees from The University
of Tokyo. He was a JSPS Fellow for Research
Abroad and a Visiting Scientist at Cornell University,
Ithaca, NY, USA, from February 2011 to February
2013. He is currently a Professor at
the Department of Information and Communication Engineering, Graduate School of Information
Science and Technology, The University of Tokyo.
His current research interests include attractiveness
computing based on multimedia big data analysis
and fundamental problems in pattern recognition and machine learning. He is
a member of ACM, AAAI, IEICE, IPSJ, and ITE.
\end{IEEEbiography}

\end{document}